\documentclass[12pt]{article}

\usepackage{lmodern}
\usepackage{amsmath,amssymb,amsthm}
\usepackage{bm}
\usepackage{mathtools}
\mathtoolsset{showonlyrefs}
\usepackage{microtype}
\usepackage{booktabs}
\usepackage[numbers,sort&compress]{natbib}
\usepackage[margin=3.6cm]{geometry}
\usepackage{authblk}
\usepackage{tabularx}

\DeclareMathOperator*{\argmax}{argmax}

\title{Deep reinforcement learning for the dynamic vehicle dispatching problem: An event-based approach}

\author{Edyvalberty Alenquer Cordeiro\footnote{Email: edyalenquer@gmail.com }}
\affil{Department of Applied Mathematics and Statistics, Federal University of Ceará, Fortaleza, Brazil}

\author{Anselmo R. Pitombeira-Neto\footnote{Email: anselmo.pitombeira@ufc.br}}
\affil{Department of Industrial Engineering, Federal University of Ceará, Fortaleza, Brazil}

\begin{document}
\maketitle

\begin{abstract}
The dynamic vehicle dispatching problem corresponds to deciding which vehicles to assign to requests that arise stochastically over time and space. It emerges in diverse areas, such as in the assignment of trucks to loads to be transported; in emergency systems; and in ride-hailing services. In this paper, we model the problem as a semi-Markov decision process, which allows us to treat time as continuous. In this setting, decision epochs coincide with discrete events whose time intervals are random. We argue that an event-based approach substantially reduces the combinatorial complexity of the decision space and overcomes other limitations of discrete-time models often proposed in the literature. In order to test our approach, we develop a new discrete-event simulator and use double deep q-learning to train our decision agents. Numerical experiments are carried out in realistic scenarios using data from New York City. We compare the policies obtained through our approach with heuristic policies often used in practice. Results show that our policies exhibit better average waiting times, cancellation rates and total service times, with reduction in average waiting times of up to 50\% relative to the other tested heuristic policies.

\end{abstract}

\section{Introduction}
\label{sec:introduction}

Taxi services play a crucial role in urban areas, which currently have the highest population densities \cite{chengServiceChoiceModel2009}. Residents of these areas have increasingly favored transportation options over purchasing their vehicles \cite{qinRideHailingOrderDispatching2020}.
The transformation of urban mobility through the use of apps such as Uber, DiDi, and Lyft has positive impacts related to sustainability and generation of income in the provision of services of this nature \cite{sadowskyImpactRideHailingServices,gaoOptimalMultitaxiDispatch2016}.

To ensure efficient service and meet demands promptly, the system must maintain continuous availability of vehicles to respond to the incoming ride requests. Additionally, minimizing waiting time is essential to prevent customer cancellations, as customers have tolerance thresholds for how long they are willing to wait. The optimization of service levels involves addressing three different problems: vehicle dispatching, vehicle relocation, and price setting \cite{qinRideHailingOrderDispatching2020}. 

In this paper, we address the dynamic vehicle dispatching problem, which corresponds to assigning vehicles to requests that arise stochastically over time and space. Previous works have treated this problem as a Markov decision process (MDP) due to its dynamic and stochastic nature. Given the MDP formulation of the problem, the goal is to approximate an optimal decision policy evaluated by a specified objective function (or loss function).

Most previous studies assume that decision epochs occur at fixed time intervals, during which the system collects information on vehicles of the fleet and pending ride requests. This information forms the state at that particular time. The decision involves determining the optimal assignment of available vehicles to waiting requests. This can be alternatively formulated as a problem of ``matching'' vehicles and requests. Although this matching-based MDP formulation has been popular in the literature, it exhibits several limitations:

\begin{enumerate}
    \item At every decision epoch, the decision space corresponds to all possible matchings of vehicles to requests. The optimal decision is ideally obtained through solving an NP-hard combinatorial optimization problem, formulated as an integer nonlinear program.  It is often reduced to a linear assignment problem due to its intractability, which can be solved efficiently but simplifies reality. (For example, the relaxed model ignores the waiting tolerance threshold of the customers, which is an important constraint in the real applications.)
    \item It is difficult to account for the possibility of both the driver and the customer rejecting the proposed ride in the mathematical programming model for the matching problem. For this reason, this matching-based approach often assumes that all proposed rides will be accepted.
    \item Since decision epochs occur at fixed time intervals, one must decide on the length of the time interval, i.e., it is a parameter that has to be previously specified or tuned by computational experimentation. Short time intervals may waste system resources while long time intervals may result in wasting valuable time that could have been used to make good assignments earlier, thus reducing customers' waiting time.
\end{enumerate}

In this paper, we seek to overcome the aforementioned limitations of the matching-based approach to the dynamic vehicle dispatching problem. In particular, our main contributions are the following:

\begin{enumerate}
    \item We propose an event-based approach for the dynamic vehicle dispatching problem that can reduce the idle time of the system and take into account some of the characteristics of the real problem that are usually disregarded in other works. More importantly, since we do not have to evaluate the set of all feasible matchings at each decision epoch, our approach can considerably minimize the complexity of the decision space.
    \item We develop a new simulation environment that uses real data to sample customers' demand and simulate the dynamics of the environment. 
    \item We develop two deep reinforcement learning agents that are our decision-makers, each being responsible for a specific event.
    \item We show that this new approach for the dynamic vehicle dispatching problem is promising by comparing the result of a large number of experiments with heuristic policies often used in practice.
\end{enumerate}

In contrast to most previous studies, which use a discrete-time approach based on the MDP framework, we develop a continuous-time approach, which is more appropriate for this problem as requests may arrive to the system and the vehicles can finish services at random times. To account for continuous time, we model the problem as a semi-Markov decision process. In this model, decision epochs are not fixed at specific time intervals but occur at random points in time. These decision epochs correspond to two types of events: when a  new request arrives in the system; or when a vehicle completes a service. This approach enables us to make decisions as soon as possible without having to specify a fixed length for the time interval between decision epochs. Additionally, since decision epochs only occur when necessary, the overall utilization of system resources is reduced.

Due to the large scale of real instances of the problem, obtaining optimal policies using exact methods is computationally infeasible, since these methods rely on enumerating all the states in the state space. To obtain good policies, we employ deep reinforcement learning and train two agents to make decisions corresponding to the two abovementioned decision events using the double deep q-learning algorithm. While our primary objective is to minimize customers' average waiting times, we also aim to avoid an increase in the cancellation rate or a decrease in drivers' income. Therefore, we simultaneously monitor these performance measures when comparing different policies.

The structure of this paper is the following: In Section \ref{sec:related}, we comment on related work on the dynamic vehicle dispatching problem; Section \ref{sec:background} revises fundamental theory on which our work is based; in Section \ref{sec:problem-formulation} we present our formulation of the dynamic vehicle dispatching problem as a semi-Markov decision process; in Section \ref{sec:proposed-solution} we detail our solution approach, which is based on reinforcement learning and simulation; Section \ref{sec:results} presents numerical results with the use of data from New York City; in Section \ref{sec:discussion} we discuss the results; finally, in Section \ref{sec:conclusions} we draw some conclusions and suggest possible future works.

\section{Related work}
\label{sec:related}
The dynamic vehicle dispatching problem (DVDP) is related with the class of dynamic fleet management and assignment problems in transportation. One of the earliest works to address a dynamic assignment problem in the context of transportation is due to \citet{powell1996stochastic}, who developed deterministic and stochastic mathematical programming models which address demand forecasts and repositioning of trucks. Approximate dynamic programming approaches to large-scale dynamic fleet management have been developed in the works of \citet{simao_powell} and \citet{godfrey_powell}. In their idea, the classical linear assignment problem is extended to a multi period environment as a discrete-time Markov decision process. At each decision epoch, many vehicles are available for assignment to many tasks (loads, in the case of trucks), resulting in a large combinatorial decision space. To cope with this, one of the strategies is to use a linear approximation to the value function, which allows one to efficiently solve the decision problem as a linear program.

In the first years of the use of digital platforms for ride hailing, companies adopted simple policies for the dispatch of vehicles to customers requests. \citet{liao2001taxi} and \citet{lee2004taxi} report on the adoption of simple myopic policies for the assignment of vehicles based on the shortest path time to arrive at a customer's location given current traffic conditions. The main disadvantage of these simple rules is that they act locally and myopically, not accounting for the global and future impact of decisions on system state, and therefore not optimizing for global performance measures. 

\citet{seow2009collaborative} formulate the DVDP at each decision epoch as a linear assignment problem with the objective of minimizing the sum of estimated current travel times from available vehicles to pending requests. Instead of solving the problem in a centralized way, the authors propose a collaborative algorithm in which groups of agents representing the drivers would negotiate and arrive at desirable assignments. The authors show through computational experiments that such approach outperforms simple rules applied by centralized systems.

\citet{zhang2017taxi} formulate an assignment problem whose objective function is the sum of acceptance probabilities by drivers. Probabilities are estimated from a logistic regression model. As the objective function of the assignment problem is nonlinear, the authors use a heuristic to solve it. \citet{bertsimas2019online} consider the ride-hailing problem as an online version of the dial-a-ride problem. They develop a re-optimization approach in which a static mixed-integer programming model is solved at fixed time intervals. The model includes all pending requests, idle vehicles and estimated travel times at the time of optimization and implement only the actions that can be applied before the next optimization epoch.

More recently, the DVDP has been addressed by using reinforcement learning (RL) techniques. \citet{xu2018large} use an MDP model in which the service provider has to match requests and available vehicles at constant time intervals. To tackle the large state space, the authors assume that the global value function is separable in local value functions associated with the drivers, and learn a value function for a single driver, whose state space is much smaller. They use past data on taxi trips to learn the value function of a base policy used during data sampling via a tabular form of dynamic programming. Decisions at real time are taken by the service provider by solving a linear assignment problem whose cost function is the sum of the learned value functions of the vehicles.

In order to overcome limitations of the tabular representation, \citet{wang2018deep} develop an approach based on deep RL in which a deep q-network receives a pair (state, action) as input and outputs the q-value associated with the pair. The network is trained by using gradient descent with data from historical trips. At runtime, the learned q-values are used to compose the objective-function of a linear assignment problem to match vehicles and requests. A similar approach is developed in the work by \citet{liang2021integrated}. \citet{tang2019deep} extend the MDP model to an SMDP and develop a new neural network architecture called ``cerebellar value networks'' to learn the value function. Most of these developments are summarized in \citet{qin2020ride}.

Other works consider the task of vehicle repositioning, either separately or integrated with order dispatching. \citet{miao2016taxi} develop a receding horizon control approach to repositioning of vehicles in anticipation of demand so as to balance the supply and demand and minimizing total idle cruising distance of vehicles.
\citet{holler2019deep} consider that decision epochs occur every time a vehicle becomes free, and then an action correspond to choose a pending request within a broadcast radius of the vehicle or reposition the vehicle in case no request is available within this radius. They train both DQNs  and proximal policy optimization to learn policies. In their results, learned policies performed comparatively better relative to simple myopic policies such as assigning the request with shortest pickup distance or with the highest revenue. 

\citet{kullman2021dynamic} have developed a deep RL approach to the DVDP with electric vehicles. At each decision epoch, a central provider must decide on which jobs to assign to each electric vehicle, such as pickup a pending request or reposition to a charging station. The authors use DQNs to learn the value function and compare the learned policies with a myopic reoptimization approach. 

\citet{liu2020context} developed a context-aware DQN-based approach to vehicle repositioning. In their work, the decision is to reposition current idle taxis to adjacent cluster locations given predicted demand in order to balance supply and demand. The matching of vehicles and requests is done by simply assigning the nearest idle vehicle. \citet{tang2021value} developed an integrated approach to dispatching and repositioning. They use a neural network to represent the state-value function and use both offline and online training. A policy corresponds to solving a linear assignment problem whose objective function is given by the sum of temporal difference errors associated with a pair (vehicle, request).

In addition to the works directly related to the DVDP summarized in this section, in the next section we revise general concepts on which we base our approach.

\section{Preliminaries}
\label{sec:background}
In this section, we revise the main theoretical concepts and ideas that underpin our work: Markov and semi-Markov decision processes, reinforcement learning  and discrete-event simulation.

\subsection{Markov and semi-Markov decision processes}

Markov decision processes (MDPs) formalize situations in which an agent (or decision maker) makes a sequence of decisions over time under uncertainty. The sequence of possible states that an agent may observe constitutes a Markov stochastic process. MDPs have been studied in connection with stochastic dynamic programming and are often used as models to sequential decision problems \cite{denardo2012dynamic, puterman2014markov}. 

\begin{figure}
\centerline{\includegraphics[width=0.85\textwidth]{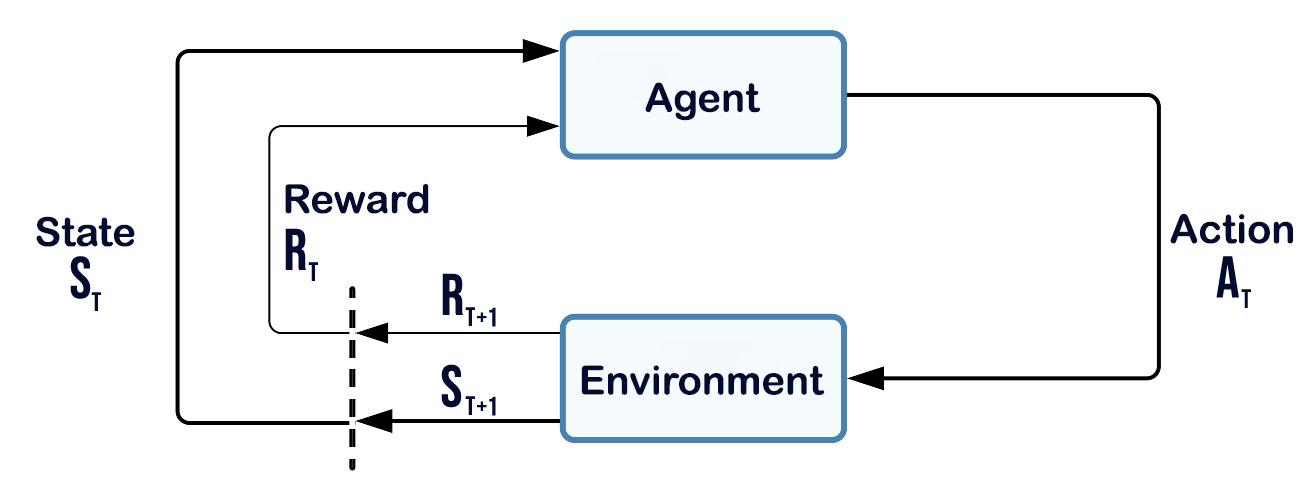}}
\caption{Representation of the dynamic interaction between MDP elements \cite{suttonBarto}.}
\label{fig:mdp-transitions} 
\end{figure}

An MDP is defined by a tuple $(\mathcal{S},\mathcal{A}, P, r, \gamma)$, in which $\mathcal{S}$ is a state set, $\mathcal{A}$ is an action set, $P$ is a probability measure, $r: \mathcal{S} \times \mathcal{A} \to \mathbb{R}$ is a reward function, and $0 < \gamma \leq 1$ is a discount factor. The state set often represents the possible states of an environment and the action set is related to an agent (a decision maker) which interacts with the environment. At a given decision epoch, the environment is in a current state $s$. The agent chooses an action $a$, receives a reward $r = r(s,a)$ and the environment makes a transition to a new state $s'$ according to a conditional probability distribution $p(s'|s,a)$. These components work together to represent the dynamics of the process, as illustrated in Figure \ref{fig:mdp-transitions}.

A solution to an MDP is called a policy, which can be regarded as a sequence of decision rules. Policies can be deterministic or stochastic. When a policy is deterministic, it always takes the same action when a given state occurs. On the other hand, when a policy is stochastic, the action taken depends on a probability distribution over all possible actions. We are interested in deterministic stationary policies, which may be defined as functions from the state set $\mathcal{S}$ to the set of actions $\mathcal{A}$.

A Markov decision problem consists in finding a policy which optimizes an objective-function. In this paper, we work with infinite-horizon MDPs, whose objective-function is
\begin{equation}
v^\star(s) = \max_{\pi \in \Pi}  \mathbb{E} \Bigg[ \sum_{t=0}^{\infty} \gamma^t r(s_t, \pi(s_t)) \Big| s_0 = s \Bigg], \quad \forall s \in \mathcal{S}, \label{eq:obj_function}
\end{equation}
$s_0$ is an initial state, $\pi: \mathcal{S} \to \mathcal{A}$ is a policy and $\Pi$ is the class of deterministic stationary policies. The function $v^\star: \mathcal{S} \to \mathbb{R}$ is known as the optimal value function. Thus, an optimal policy $\pi^\star$ maximizes the objective-function \eqref{eq:obj_function} for all initial states.

The optimal value function $v^\star$ can be obtained by solving the \emph{Bellman equation} \cite{bellmanDynamicProgramming1957}:

\begin{equation}
v^\star(s)  = \max_{a \in \mathcal{A}_{s}} \{r(s, a)+\gamma \mathbb{E}[v^\star(s')]\},
	    	 \quad \forall s \in \mathcal{S},
\end{equation}
in which  $\mathcal{A}_s \subseteq \mathcal{A}$ is the set of feasible actions at state $s$. It can be shown that knowledge of $v^\star$ leads directly to an optimal policy by simply acting greedily in relation to the the optimal value function:
\begin{equation}
\pi^\star(s) \in \arg\max_{a \in \mathcal{A}_{s}} \{r(s, a)+\gamma \mathbb{E}[v^\star(s')]\},
	    	 \quad \forall s \in \mathcal{S}.
\end{equation}

There are three main exact methods to obtain an optimal policy: value iteration \cite{bellmanDynamicProgramming1957}, policy iteration \cite{howard1960dynamic} and linear programming \cite{d1963probabilistic, manne1960linear}. These methods have been applied to a variety of problems, such as robotics control \cite{levine2016end}, and game playing \cite{mnihHumanlevelControlDeep2015}. Value Iteration uses the Bellman equation to compute the optimal value function iteratively. Policy iteration combines policy evaluation and policy improvement steps, where the Bellman equation is used to evaluate the policy. 

Semi-Markov decision processes (SMDPs) generalize MDPs by allowing decision epochs to occur at deterministic irregular time intervals or random time intervals \cite{suttonMDPsSemiMDPsFramework1999,bartoRecentAdvancesHierarchical,bertsekas2012dynamic}. SMDPs extend the class of problems which can be modeled by the MDP formalism as the following:
\begin{enumerate}
    \item Decision epochs do not need to occur at every state transition and may be started by events corresponding to so called \emph{decision states}.
    \item Rewards may be a function of the time interval length.
    \item The environment may remain at a given state for a variable time interval.
\end{enumerate}
SMDPs also allow the decoupling of the \emph{natural process} and the \emph{decision process}. The natural process corresponds to the sequence of all states of the environment over time, while the decision process corresponds to the sequence of only the decision states. In the MDP formalism these two processes coincide.

Since we assume continuous time in SMDPs, the objective function must be modified as follows:
\begin{equation}
v^\star(s) = \max_{\pi \in \Pi}  \mathbb{E} \Bigg[ \sum_{k=0}^\infty \int_{t_{k}}^{t_{k+1}} e^{-\beta t} r(s_k, \pi(s_k)) dt \Big| s_0 = s \Bigg], \quad \forall s \in \mathcal{S}, \label{eq:obj_function_smdp}
\end{equation}
in which $s_k$ is the state at decision epoch $t_k$. Notice in \eqref{eq:obj_function_smdp} that the time interval between consecutive decision epochs $t_k$ and $t_{k+1}$ is a random variable, so that we must integrate the reward function over the interval and $e^{-\beta t}$ is the limiting form of the discrete-time discount factor $\gamma$ for continuous time, with $e^{-\beta} = \gamma$.

Let $R(s,a)$ be the expected reward between two consecutive decision epochs:
\begin{equation}
    R(s,a) = \mathbb{E}\Big[\int_0^{\tau} e^{-\beta t}r(s,a)dt\Big]\Big\},
\end{equation}
in which the expectation is taken relative to the probability distribution of soujourn times $\tau$ between decision epochs. Notice that, in contrast to MDPs, in SMDPs the rewards depend on the random time intervals between decision epochs.

The optimally Bellman equation for an SMDP can be written as
\begin{equation}
v^\star(s)  = \max_{a \in \mathcal{A}_{s}} \Big\{R(s,a)+\mathbb{E}\Big[\int_0^{\infty} e^{-\beta t}v^\star(s')dt\Big]\Big\},
	    	 \quad \forall s \in \mathcal{S},
\end{equation}
and the expectation is taken relative to the distribution of next states $s'$. 

\subsection{Reinforcement learning}

Reinforcement learning (RL) has emerged as a collection of solution methods within the artificial intelligence community, specifically designed to address problems concerning the training of artificial agents in decision-making over time. RL theory encompasses the employment of Markov Decision Processes (MDP) and Semi-Markov Decision Processes (SMDP) as general mathematical models for solving RL problems. In recent years, RL has garnered considerable attention and acclaim due to its remarkable achievements in domains such as board games \cite{silverGo}, robotics \cite{RL_Robotica, liu2023safe}, video games \cite{mnih2015human}, and operations research \cite{pitombeira2022reinforcement,SAMSONOV2022,JIN2023}. RL algorithms exhibit several similarities to approximate dynamic programming techniques \cite{powell_book, bertsekasNeuro} in various aspects.

RL is particularly valuable when dealing with sequential decision problems where the underlying model is unknown. To address such scenarios, RL employs approximation techniques. In many real-world applications, such as robotics or game playing, it is challenging, if not impossible, to model the environment and the consequences of actions precisely. In these cases, RL provides a framework for learning a policy or a value function that can guide the decision-making process based on experience gathered through trial-and-error interactions with the environment. RL makes use of Q-functions to overcome the need for a model:
\begin{equation}
    q^\star(s,a) = R(s,a)+\mathbb{E}\Big[\int_0^{\infty} e^{-\beta t}v^\star(s')dt\Big],\,
	    	 \quad \forall s \in \mathcal{S}, \forall a \in \mathcal{A}.
\end{equation}
Q-functions satisfy a form of Bellman equation:
\begin{equation}
    q^\star(s,a) = R(s,a)+\mathbb{E}\Big[\int_0^{\infty} e^{-\beta t} \max_{a' \in \mathcal{A}_{s'}}q^\star(s',a')dt\Big],
	    	 \quad \forall s \in \mathcal{S}, \forall a \in \mathcal{A}.
\end{equation}

Q-functions allow us to apply a form of online value iteration, known as q-learning, which needs only samples of state transitions, actions and rewards. The sampled transitions may be obtained from a real system or through simulation. Given a sample transition $(s, a, r, \tau, s')$, q-learning updates the value $q(s,a)$ by \cite{bradtke1994reinforcement};
\begin{equation}
    q(s,a)  \gets q(s,a)+\alpha \Big(r+e^{-\beta \tau} \max_{a' \in \mathcal{A}_{s'}}q^\star(s',a') - q(s,a)\Big),
    \label{eq:q-learning-update}
\end{equation}
in which $r$ is the observed time-dependent reward, $\tau$ is the observed time between consecutive decision epochs, and $ 0 < \alpha \leq 1$ is a learning rate.

One of the primary challenges in RL is to handle the curse of dimensionality, which arises due to the exponential growth in the number of possible states and actions. To overcome this issue, RL algorithms often rely on function approximation methods, such as neural networks or decision trees, to approximate the optimal value function or policy. 

\begin{figure}
\centerline{\includegraphics[width=0.9\textwidth]{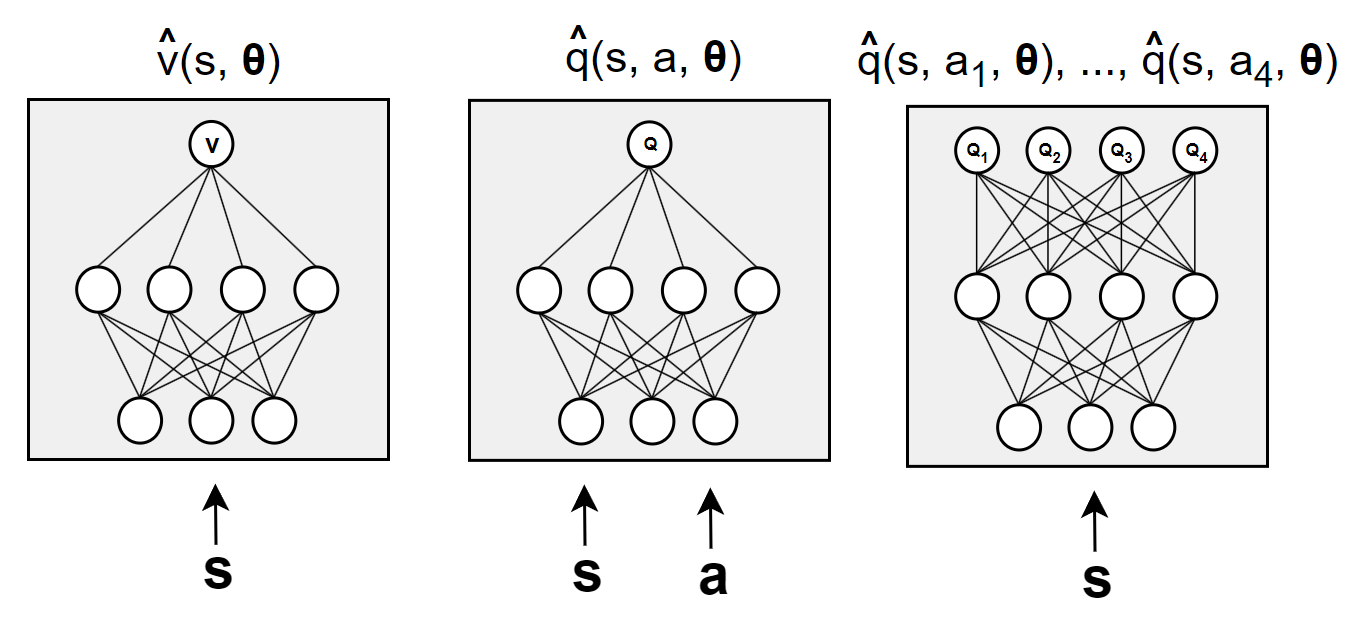}}
\caption{Three different ways in which a neural network may be used to approximate a value function \cite{silver2015}.}
\label{fig:aprox-aproaches}
\end{figure}

An approximate model for the optimal q-function is denoted by $\hat{q}(s,a;\bm{\theta})$, in which $\bm{\theta}$ is a set of parameters that must be trained on sampled data. Fig \ref{fig:aprox-aproaches} illustrates how we can use a neural network to approximate a value function.

In the leftmost side in Fig. \ref{fig:aprox-aproaches}, the approximation involves providing the state as input to a neural network, which outputs the state-value for the given state. In the center form, both the state and action are given as input to the network, and the output is the action-value for the given state and action pair. Finally, in the rightmost form, only the state is provided as input, and the output is the action-value for all feasible actions given the state.

\subsection{Discrete-event simulation}
In order to train an agent, we must obtain samples from state transitions, actions and rewards. Although in principle we can use samples collected from interactions of the training agent with a real environment, this is generally inefficient and risky. Samples are often obtained by computer simulation. In this work, we obtain samples by using discrete-event simulation (DES).

DES is a powerful simulation paradigm which allows us to simulate discrete-state systems which evolve in continuous time \cite{law2014simulation, gosaviSimulationBasedOptimization2015, ross2022simulation}. DES has been successfully applied to queuing systems, transportation systems \cite{reis2017simulation} and discrete-event dynamic systems in general. 

DES exploits the fact that stochastic discrete-state systems remain at the same state for random finite time intervals and state transitions occur only at discrete points in time (events). A DES simulator keeps a record of random events in a list and jumps between events, updating system states and collecting statistics only at the times in which events occur. This greatly accelerates simulation and makes efficient use of computational resources.

\section{Formulation of the dynamic vehicle dispatching problem as a semi-Markov decision process}
\label{sec:problem-formulation}

In the following paragraphs, we formulate the process of dynamically dispatching vehicles to customers requests as a semi-Markov decision process. As such, we define the basic components of an SMDP: the environment, the state  and action spaces, the reward function, and the environment dynamics.

\subsection{Environment representation}

\begin{figure}
\centerline{\includegraphics[width=0.7\textwidth]{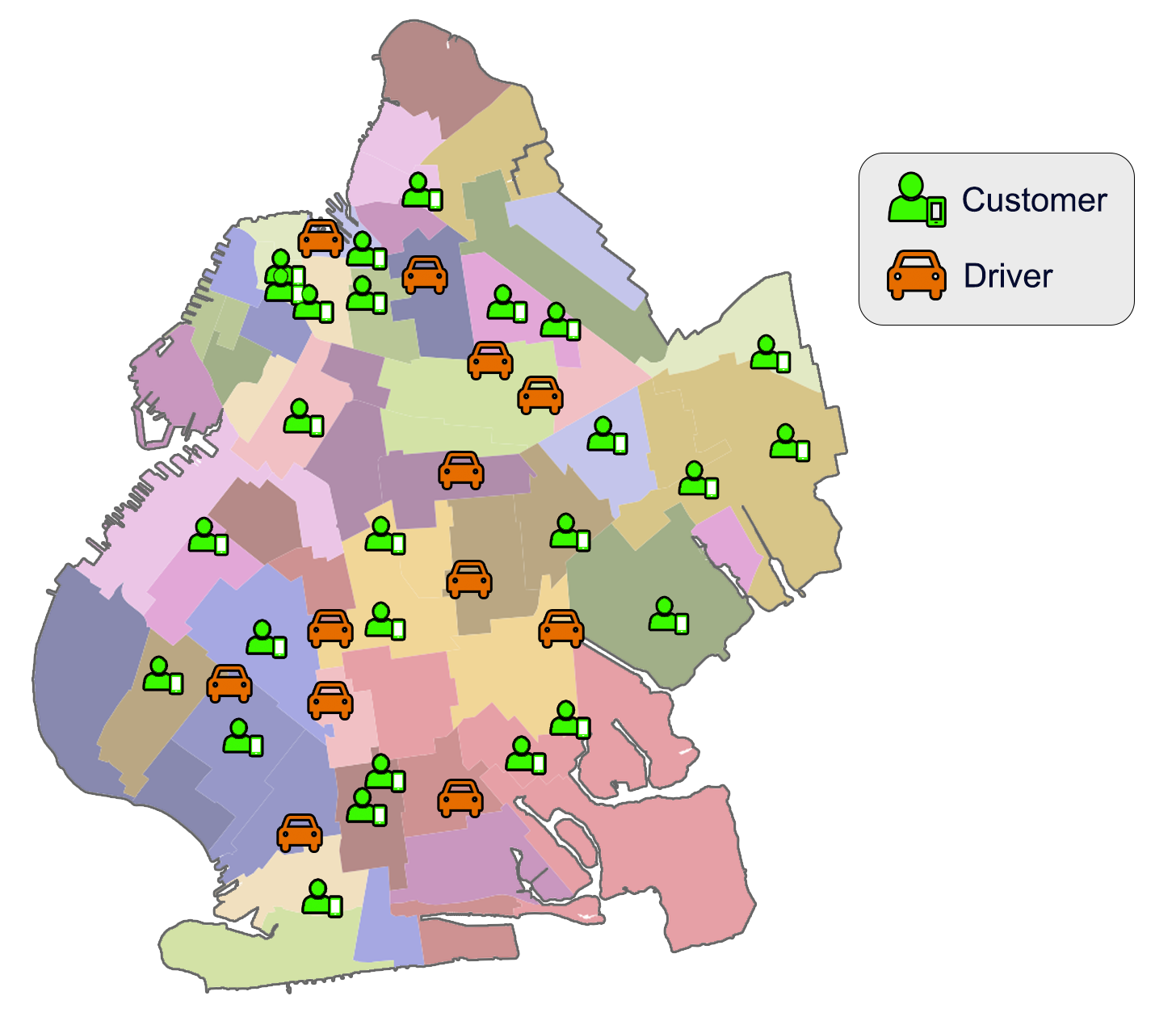}}
\caption{Spatial representation of the environment in the dynamic vehicle dispatching problem}
\label{fig:space-representation-mdp}
\end{figure}

We represent the environment as the space where trip requests can originate, and we use a simple coordinate system based on latitude and longitude to represent the location of vehicles and trips.  Fig. \ref{fig:space-representation-mdp} illustrates an example environment with a spatial distribution of vehicles and requests at a given time. It is important to notice that our model assumes complete knowledge of the environment, allowing the system to utilize information about all vehicles and requests at any time, such as locations and current status. Due to the high degree of digitalization of current vehicle dispatching systems, this is a realistic assumption.

\subsection{Disadvantages of an MDP formulation}

\begin{figure}
\centerline{\includegraphics[width=0.9\textwidth]{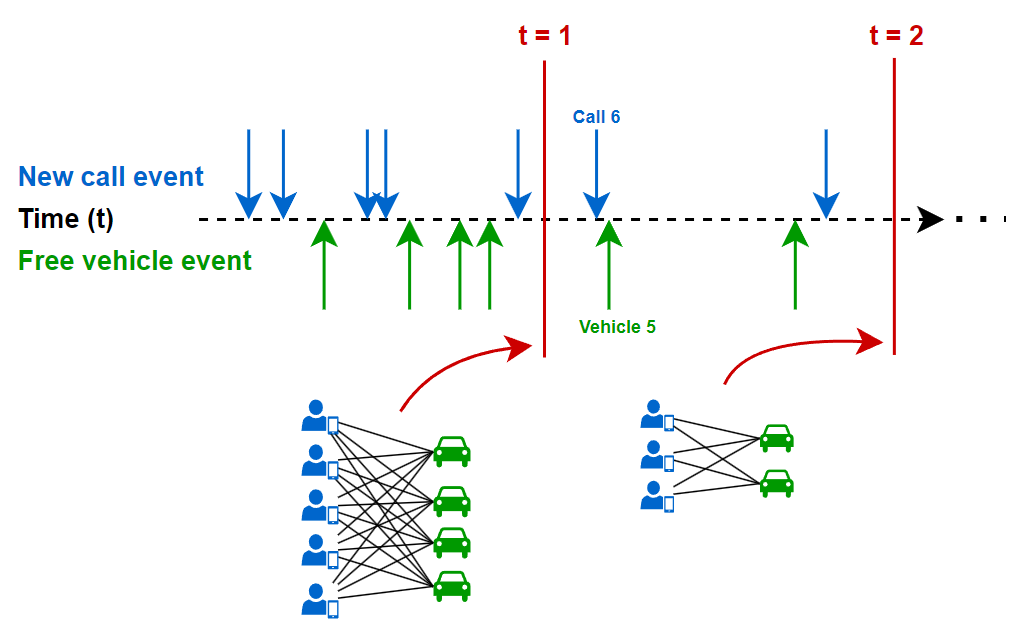}}
\caption{Decision epochs in the dynamic vehicle dispatching problem modeled as an MDP. Since decision epochs occur at fixed time intervals, the agent has to solve an NP-hard generalized assignment problem at each decision epoch.}
\label{fig:mdp}
\end{figure}

Most previous works in the literature model the DVDP as an MDP. At discrete-time decision epochs, a decision agent (e.g., a mobility software platform) observes the current state of the system and has to decide which vehicles will serve which requests. This involves solving a matching problem, whose set of possible solutions may be very large due to its combinatorial nature. Fig. \ref{fig:mdp} shows an example of how the DVDP is solved using a matching approach with an MDP framework. The time intervals between decision epochs in the MDP are of equal length, which can be a limitation in solving certain problems. For the DVDP, this characteristic of MDPs can result in two significant consequences. First, the system must wait for a fixed amount of time between consecutive decision epochs, which can result in multiple events occurring before the next decision epoch is triggered. 

For example, at time $t=1$ in Fig. \ref{fig:mdp}, the system has $n = 5$ ride requests waiting and $m = 4$ vehicles available. In this case, there are $5! = 120$ possible matchings (assuming  the existence of one dummy vehicle). In real-world problems, the number of possible solutions is $O(\max(n,m)!)$, making it hard to find the best (or even a good) decision in a timely manner. Finding the best matching corresponds to solving a generalized assignment problem, which is known to be NP-hard. The approaches based on MDP circumvent this difficulty by relaxing the full problem to a linear assignment problem, which can be solved efficiently, but ignores many constraints of the real problem. 

A second implication of the MDP formulation is that the fixed time intervals between decision epochs also affect customers' waiting time. In the example depicted in Fig. \ref{fig:mdp}, when vehicle 5 becomes free, there are 2 ride requests waiting for assignment. These ride requests need to wait until the second decision epoch at $t=2$ before they can potentially receive an assignment. This can result in an increase of waiting time since vehicle 5 could have been assigned to request 6 as soon as it got free.

In summary, in real-world scenarios, the environment can be very large, and making optimal decisions can become very hard when using an MDP formulation.

\subsection{Proposed event-based SMDP formulation}
\label{sec:SMDP_formulation}

We notice that we can overcome many of the abovementioned disadvantages of the MDP formulation by adopting an event-based approach, which reduces the complexity of the assignment problem involved. We define two important events in the system, which trigger decision epochs: the rise of a new call (new call event) and the completion of a vehicle trip (new vehicle event). 

\begin{figure}
\centerline{\includegraphics[width=0.7\textwidth]{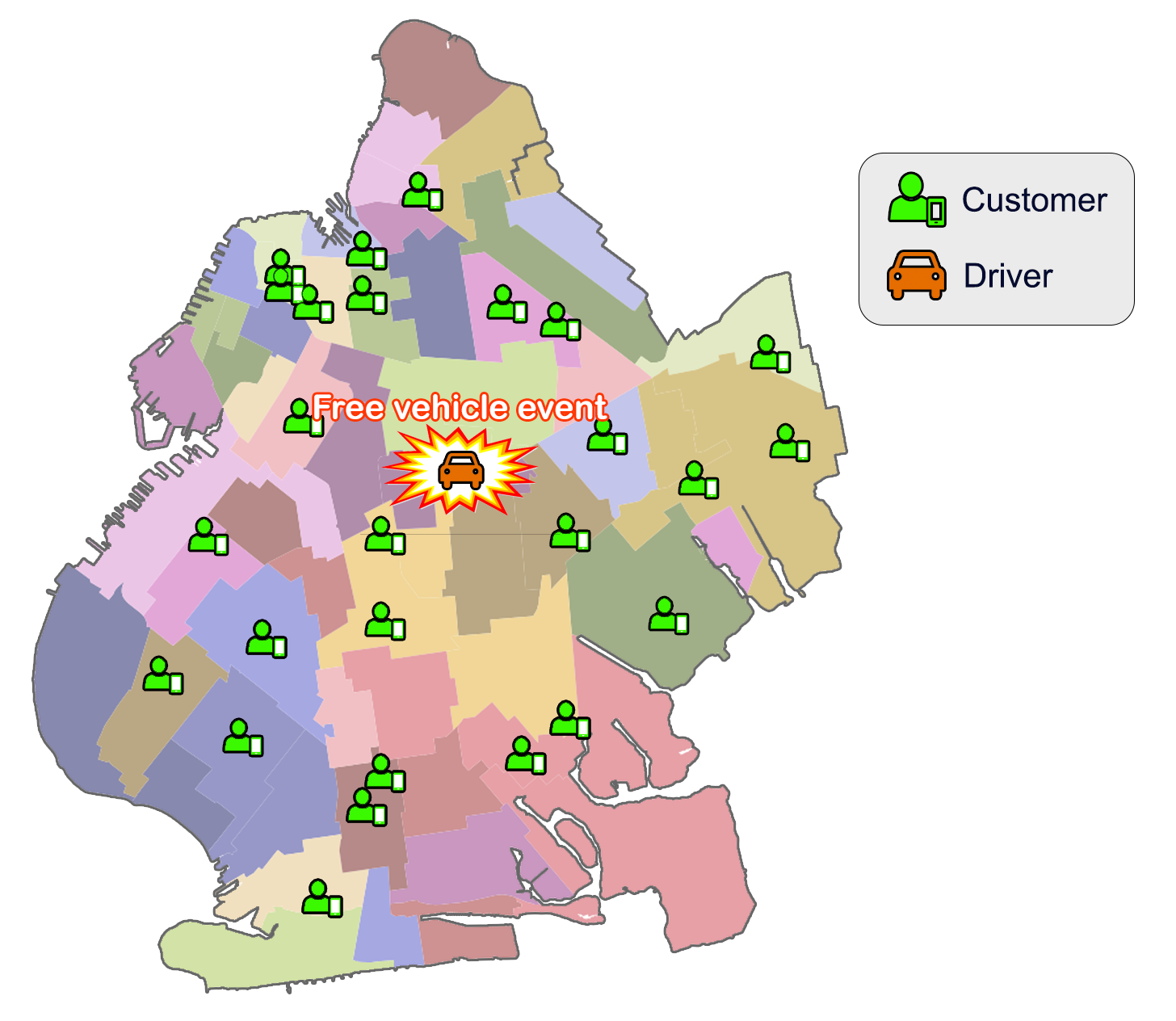}}
\caption{Spatial representation of a free vehicle event. All shown customers are waiting for service.}
\label{fig:spatial-repres-free-vehicle}
\end{figure}

\begin{figure}
\centerline{\includegraphics[width=0.7\textwidth]{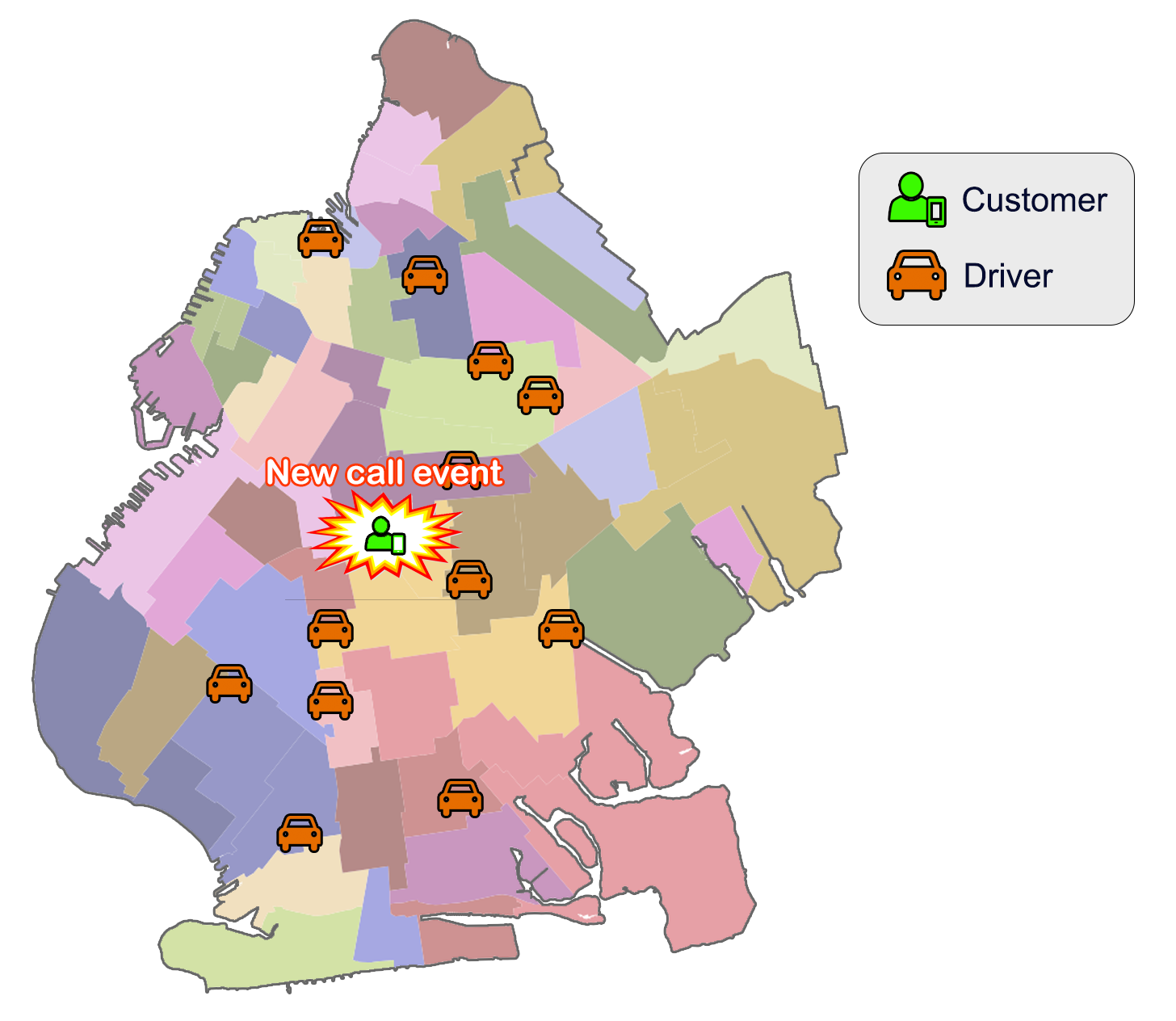}}
\caption{Spatial representation of new call event, in which one of the availables cars must be assigned to the call.}
\label{fig:spatial-repres-new-call}
\end{figure}

Figure \ref{fig:spatial-repres-free-vehicle} illustrates the free vehicle event. Notice that, in this case, we only need to decide which request to assign the vehicle from the pool of waiting requests. This assumes there are no other vehicles available at this same time, since in a scenario with more waiting requests than available vehicles, a free vehicle must have been assigned in a decision epoch before the current one. This results in a one-to-many assignment problem, which is simpler than the many-to-many assignment problems encountered when using an MDP approach. The same reduction in assignment complexity occurs when a new call event arises in the system, as illustrated in Figure \ref{fig:spatial-repres-new-call}. In this case, the problem is reduced to selecting the best vehicle to serve the new request.

\begin{figure}
\centerline{\includegraphics[width=0.9\textwidth]{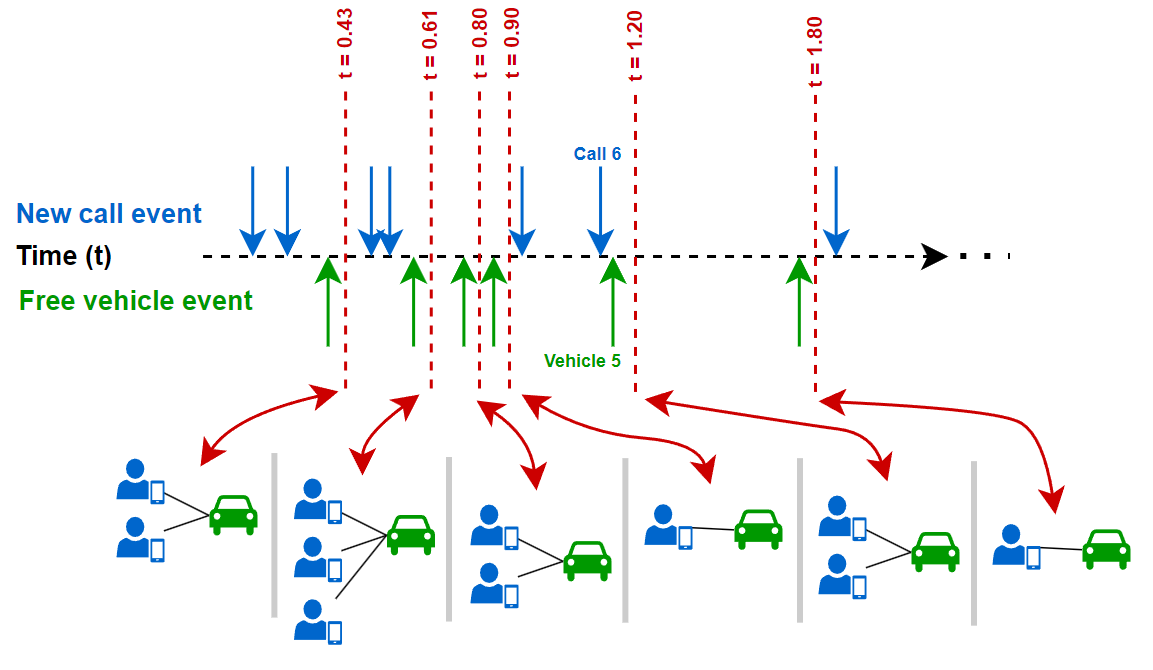}}
\caption{Decision epochs in the dynamic vehicle dispatching problem modeled as an MDP. Since decision epochs occur as soon as an event occurs, the agent solves a simple one-to-many assignment problem.}
\label{fig:smdp}
\end{figure}

Since the events occur stochastically over time, the time intervals between consecutive decision epochs are  random. This requires us to model the DVDP as an SMDP. The first advantage of this formulation is that decisions are made as soon as possible, with no need to wait for a decision epoch at fixed time intervals. This also eliminates the problem of determining the length of the fixed time intervals between decision epochs. The second advantage is the reduction in decision complexity, which consistently transforms assignment problems into one-to-many configurations, as mentioned earlier because we focus on the entity that triggers the observed event. Fig. \ref{fig:smdp} shows how the decision epochs occur over time in our approach.

In Fig. \ref{fig:smdp}, it is important to notice that the decision-making epochs, represented by the red dotted vertical lines, occur at irregular intervals in continuous time. By anticipating the decision-making moment, the system has the opportunity to establish a good assignment whenever possible, in contrast to the traditional MDP approach that requires waiting for a predetermined moment to make a decision. In particular, notice that vehicle 5 can be assigned to call 5 or call 6 as soon as it gets free (cf. Fig. \ref{fig:mdp}), reducing their waiting times. Regarding the increase of decision epochs, this will not be a problem, as the assignment problems in this approach are substantially simpler and are solved quickly.

\subsection{State and action spaces}

At each decision epoch, the system state faced by the agent depends on the event triggered. For instance, in the event of a free vehicle, the state will be characterized by the specific vehicle, while the actions to be taken will be determined based on the pending requests. Conversely, when a new request (a new call) event occurs, the state will be defined in terms of the request, and the actions will be established by considering the available vehicles within the fleet. This adaptive approach ensures that the data structure provided to the agents aligns with the specific event and enables them to make informed decisions based on the pertinent information.

Table \ref{tab:vehicle-state-repr} presents the seven features related to the vehicles. When the free vehicle event is triggered, these features represent the state. When the new call event is triggered, the state refers to the trip that arrives in the system. Table \ref{tab:trip-state-repr} shows the features related to incoming calls. Furthermore, in both data structures, we append some context features. Table \ref{tab:context-state-repr} presents the three context features considered in this work.
\begin{table}[t]	
\centering
    \begin{tabularx}{\textwidth}{>{\hsize=.1\hsize}X>{\hsize=.25\hsize}X>{\hsize=.65\hsize}X}
        \toprule
        Index & Feature & Description\\
        \midrule
        1 & X coordinate of the vehicle location & The X coordinate of the current vehicle location. \\
        2 & Y coordinate of the vehicle location & The Y coordinate of the current vehicle location. \\
        3 & X coordinate of the vehicle destination & 
        The X coordinate corresponds to the destination of the vehicle's movement. In the case where the vehicle is not occupied, this coordinate is set to the same coordinate of the index 1. \\
        4 & Y coordinate of the vehicle destination & The Y coordinate corresponds to the destination of the vehicle's movement. In the case where the vehicle is not occupied, this coordinate is set to the same coordinate of the index 2. \\
        5 & Time to finish the service & Estimated time for the vehicle arrive at the destination of the call. In the case where the vehicle is not occupied, this feature is set to $0$. \\
        6 & Cancel probability & The probability of the driver reject an assignment suggested by the system. \\
        7 & Vehicle status & This feature is set to $0$ when the vehicle is idle and $1$ when the vehicle is occupied. \\
        \bottomrule
    \end{tabularx}
	\caption{\label{tab:vehicle-state-repr} Vehicle state features}	
\end{table}
\begin{table}	
	\centering
	
        \begin{tabularx}{\textwidth}{>{\hsize=.1\hsize}X>{\hsize=.25\hsize}X>{\hsize=.65\hsize}X}
			\toprule
			Index & Feature & Description\\
			\midrule
			1 & X coordinate of the call origin location & The X coordinate of the call origin location. \\
			2 & Y coordinate of the call origin location & The Y coordinate of the call origin location. \\
			3 & X coordinate of the call destination location & The X coordinate of the call destination location.  \\
			4 & Y coordinate of the call destination location & The Y coordinate of the call destination location.  \\
			5 & Time the ride request arrived in the system & The time that the customer sent the ride request to the system. \\
            \bottomrule
		\end{tabularx}	
	\caption{\label{tab:trip-state-repr} Ride request features}	
\end{table}
\begin{table}	
	\centering
    \begin{tabularx}{\textwidth}{>{\hsize=.1\hsize}X>{\hsize=.25\hsize}X>{\hsize=.65\hsize}X}
        \toprule
        Index & Feature & Description\\
        \midrule
        1 & Resource/demand ratio & Quantity of vehicle of the fleet divided by the quantity of new calls that arrives at the last 15 minutes. At the beginning of the process this is set to 1. \\
        2 & Week cyclical feature 1 & Week time encoded with the following formula: $$\sin(2\pi( m / 10080)) $$ where $m$ represents the minute of the week.\\
        3 & Week cyclical feature 2 & Week time encoded with the following formula: $$\cos(2\pi( m / 10080)) $$ where $m$ represents the minute of the week.\\
        \bottomrule
    \end{tabularx}
	\caption{\label{tab:context-state-repr} Context features}
\end{table}

The action space also depends on the type of event. In the case of a free vehicle event, the action space corresponds to the set of all waiting requests. The agent then chooses which request will be served by the current free vehicle (See Fig. \ref{fig:spatial-repres-free-vehicle}). In the case of a new call event, the action space corresponds to the set of all currently available vehicles. The agent then chooses which vehicle will serve the new request (See Fig. \ref{fig:spatial-repres-new-call}). We emphasize the reduction of the action space, which encompasses only a one-to-many assignment, while an MDP formulation demands a hard many-to-many assignment. 

\subsection{Reward function}
\begin{figure}
\centerline{\includegraphics[width=0.9\textwidth]{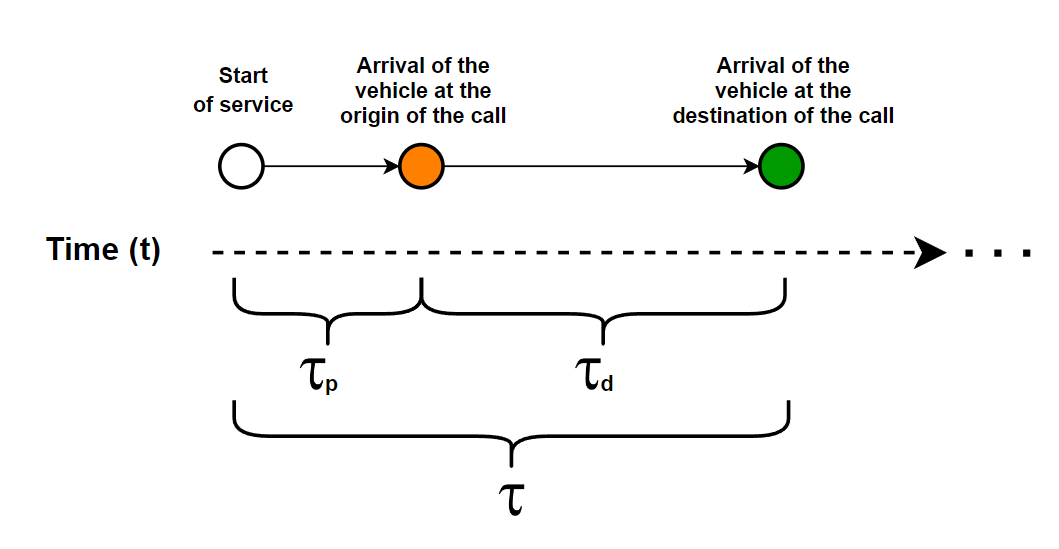}}
\caption{Times involved in the call handling process}
\label{fig:process}
\end{figure}

The agents in this work receive rewards based on the duration of the trip and customer waiting time. The time segmentation of the service process used in this work is presented in Fig. \ref{fig:process}.  We denote the total service time as $\tau$, which corresponds to the time interval between the assignment of a vehicle to a call and the arrival of the vehicle at the ride destination. $\tau$ is the sum of $\tau_p$, the time interval between vehicle assignment and its arrival at the request origin, and $\tau_d$, the time until the arrival at the ride destination.

The plain reward $R$ associated with the agent assigning a vehicle to a request is defined as proportional to the estimated time of the ride:
\begin{alignat}{3}
    \label{eq:undiscounted-reward}
    R = \hat{\tau}_{d} + b,
\end{alignat}
in which $\hat{\tau}_{d}$ represents the estimated time it takes to travel from the origin to the destination of the ride request. The idea is that longer rides will provide higher monetary rewards to the drivers. 

However, we also want the agent to increase the rates of served requests, which will also indirectly reduce the waiting times and cancellation rates. We then added a constant term $b$, which acts as a fixed reward independent of the estimated duration of the ride $\hat{\tau}_{d}$. The parameter $b$ represents a hyperparameter in our model that requires specification based on empirical data obtained from real-world applications. For instance, within the context of a ride-hailing platform, $b$ may denote a fixed fare charged for each trip or can be determined through computational optimization. It is worth noting that a higher value of $b$ encourages agents to undertake additional trips, thereby indirectly reducing the overall duration of individual trips. Conversely, a lower value of $b$ provides incentives for agents to opt for longer trips, aiming to maximize the cumulative reward obtained.

In addition, due to the time-dependence of the problem, the reward signal that the agents receive is further discounted over time. For this, we use the accumulated discounted reward:
\begin{alignat}{3}
    R_{disc} &  = \dfrac{R}{\tau} + \gamma \dfrac{R}{\tau} + \gamma^{2} \dfrac{R}{\tau} + \cdots + \gamma^{\tau-1} \dfrac{R}{\tau}\\
    & = \dfrac{R(\gamma^{\tau} - 1)}{\tau(\gamma - 1)},
    \label{eq:discounted-reward}
\end{alignat}
where $R$ is the plain reward, $\gamma$ the discount factor and $\tau$ is the total service time, as shown in Fig. \ref{fig:process}.

\subsection{Environment dynamics}
The environment dynamics evolves continuously over time. State transitions occur at discrete points in time. Thus, the environment may remain at a given state for a random amount of time. For example, position of vehicles make part of the state of the system, and we assume these change only at discrete points in time, such as when a vehicle arrives at a request's origin or destination.

The environment encompasses four main sources of randomness: Ride requests appear in the environment at random times and locations, the movement of vehicles between locations takes varying durations, drivers may refuse to serve a request offered by the decision agent with a given probability, and customers have an unknown probability distribution associated with their maximum tolerance time to wait.

The logic of the environment dynamics is embedded in a discrete-event simulation model, whose details are described in Section \ref{sec:simulator}.

\section{Proposed solution approach}
\label{sec:proposed-solution}

Given the SMDP formulation of the DVDP, we could in theory use exact methods such as value iteration or policy iteration to obtain an optimal decision policy. However, the state space in the DVDP is too large and exact methods rely on enumerating all states, which makes exact methods computationally infeasible (the well known \emph{curse of dimensionality}). Moreover, since the environment is very complex, we also do not have direct access to the conditional probability distributions of state transitions and can only sample the environment dynamics through a simulator.

To address these issues, we use model-free reinforcement learning techniques. We approximate the q-functions using deep neural networks which are trained from experiences sampled from the discrete-event simulator and use the double q-learning algorithm to adjust the weights of the networks. The subsequent sections outline our proposed solution approach.

\subsection{General training scheme}

Since in our SMDP formulation of the DVDP, decision epochs occur at two different types of events, we train two different decision agents: the \texttt{NewCallAgent} and the \texttt{FreeVehicleAgent}. We train our agents by combining real-world and simulated data. Our simulator utilizes the origin and destination locations and the arrival times of trips from  real data. Additionally, we generate some simulated data using probability distributions, which will be discussed in the following sections. Our agents make decisions based on q-value estimates, which measure the quality of a given action in a specific state. The q-values are the output of deep neural networks. 

\begin{figure}[t]
\centerline{\includegraphics[width=0.9\textwidth]{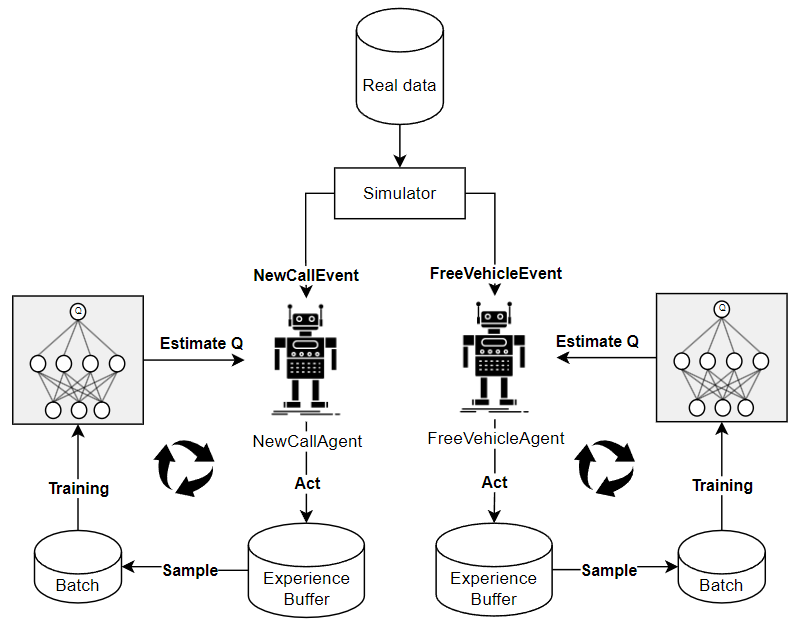}}
\caption{General training scheme of the two decision agents}
\label{fig:general-scheme}
\end{figure}

During the initial stage of the training process, the agents tend to make random decisions as they learn about the consequences of their actions in the environment. A sample transition is represented by a tuple 
\[\text{(current state, next state, action, reward, event duration)}.\] As the simulation goes on, sample transitions are stored in a pool called the ``experience buffer'' (EB). Each agent has its neural network. To train these neural networks, we randomly sample a batch of experience tuples from the EB at every successful assignment and optimize their weights using the double Q-learning algorithm. Due to the random occurrence of decision epochs and the eventual subsequent assignments, the agents are trained simultaneously during the simulation process, without a predefined order. This general scheme is presented in Fig \ref{fig:general-scheme}.

\subsection{Discrete-event simulator}
\label{sec:simulator}

To implement the simulation environment, we utilize DES, which aligns well with our event-based approach. We implement the simulator with the aid of SimPy, a DES library for the Python language.

\subsubsection{Entities, agents and environment}
\begin{figure}
\centerline{\includegraphics[width=0.9\textwidth]{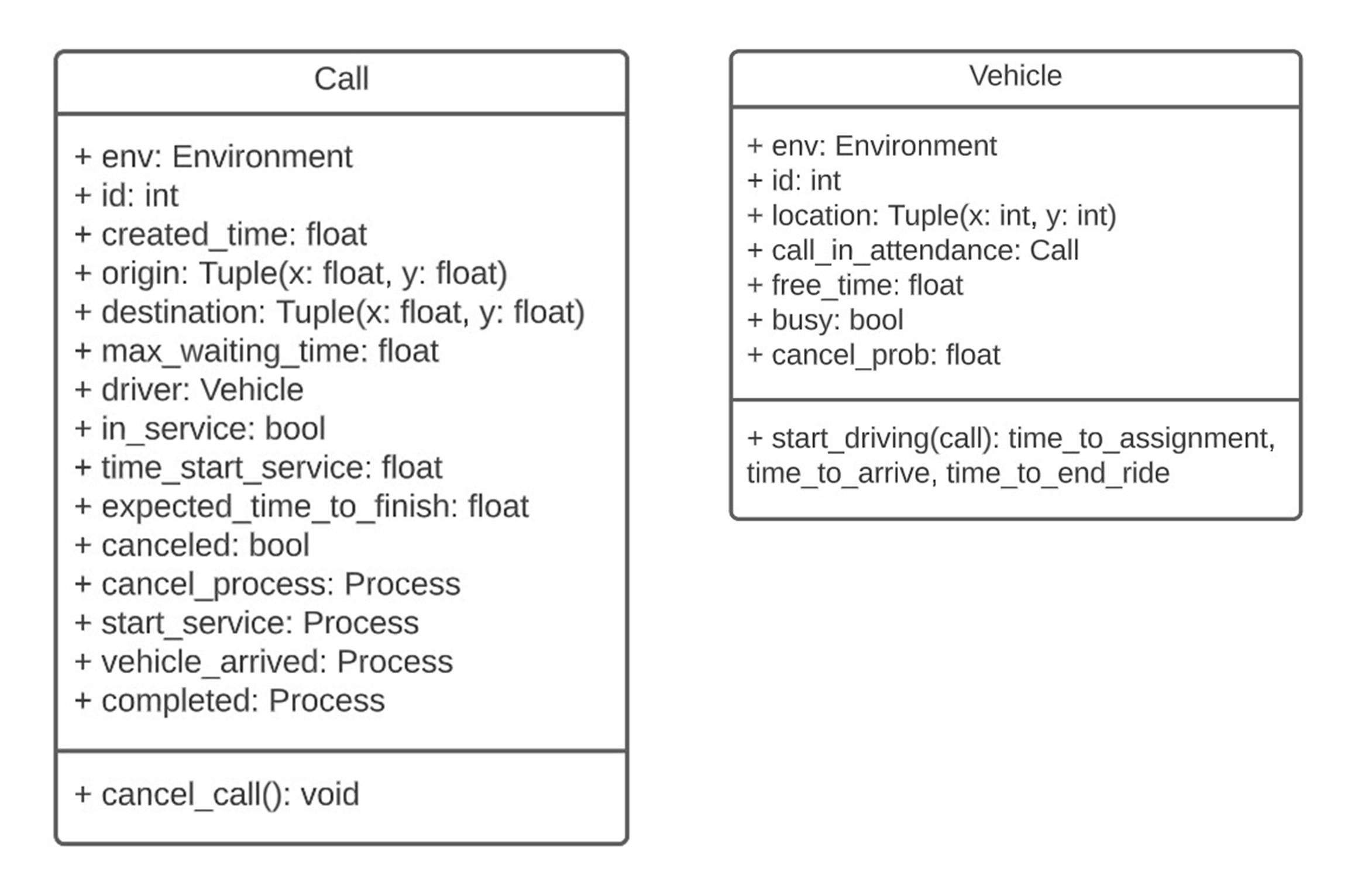}}
\caption{Class definitions of the basic entities call and vehicle. Squares in the top show class attributes while squares in the bottom show class methods.}
\label{fig:basic-entities}
\end{figure}

We define three categories of objects which compose the simulator: basic entities, agents (decision makers) and the environment. The two diagrams displayed in Fig. \ref{fig:basic-entities} show the class definitions of the basic entities: calls and vehicles.

The \texttt{Call} class represents the ride requests made by customers. The main attributes are the time of creation, origin, destination, and maximum waiting time the customer is willing to wait to be served. \texttt{Call} objects are created during the course of a simulation by a function that based on real-world data draws their creation times, origins, and destinations and randomly defines the maximum waiting time.

The \texttt{cancel\_call()} method of the \texttt{Call} class is a process function as defined in the SimPy library. Process functions  are implemented as coroutines, which are special kinds of functions that can be interrupted and resumed at a later time at the same point where it stopped executing. The \texttt{cancel\_call()} method is started when a new object of the \texttt{Call} class is instantiated, and its objective is to monitor the maximum time that each specific customer can tolerate to wait. When the tolerance is reached, the \texttt{cancel\_call()} method removes the call from the waiting calls pool. 

The \texttt{Vehicle} class represent the vehicles in the simulator. The main attributes of a vehicle are its current location and busy status. Each vehicle is created at the start of a simulation and may serve many calls during a simulation run. The \texttt{start\_driving()} method of the \texttt{Vehicle} class is also a 
 SimPy process function. Is starts when a vehicle is assigned to a call and finishes when the vehicle arrives at the destination, when it triggers a \texttt{FreeVehicleEvent}.
 \begin{figure}
\centerline{\includegraphics[width=0.85\textwidth]{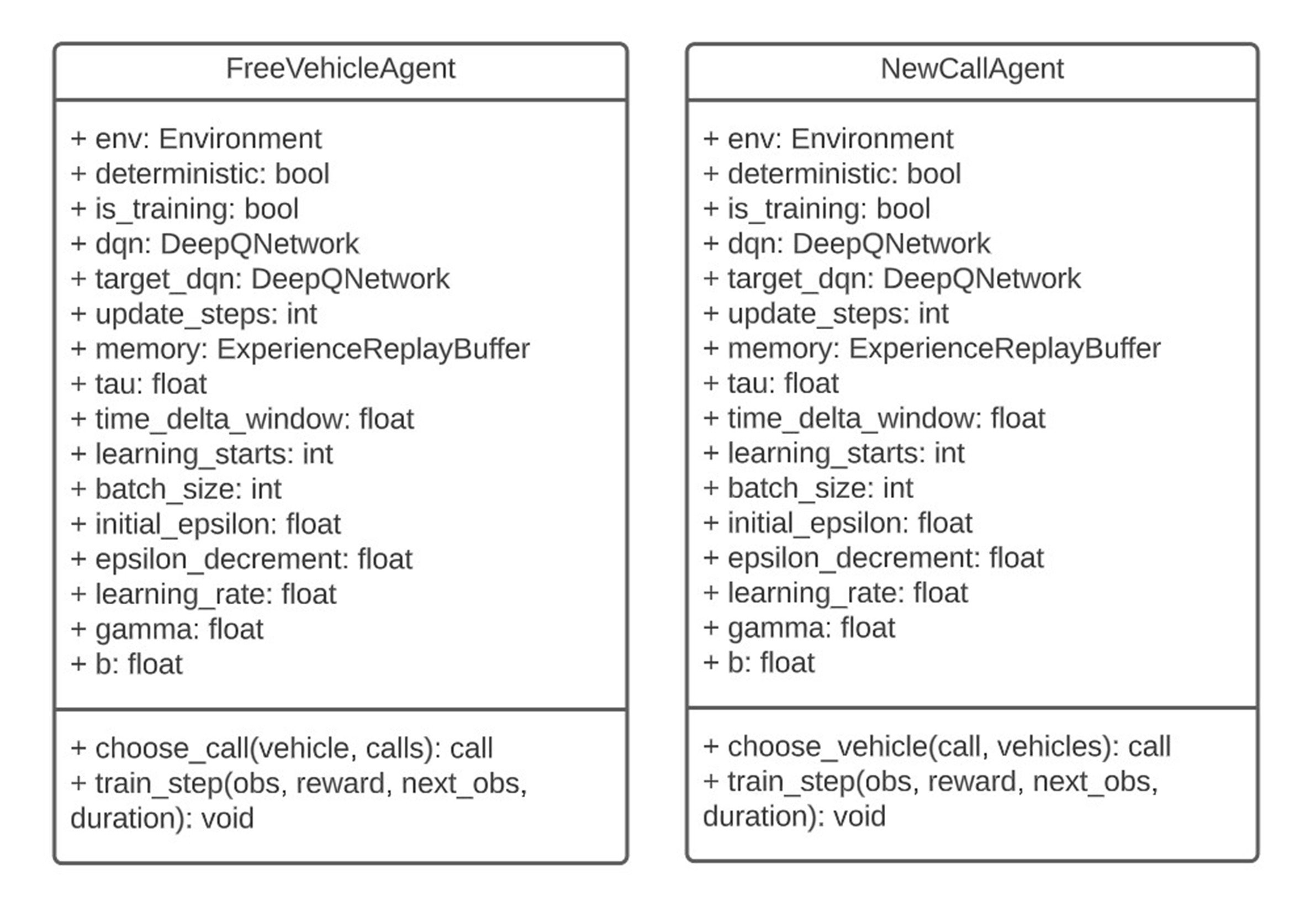}}
\caption{Class definitions of the two decision agents. Squares in the top show class attributes while squares in the bottom show class methods.}
\label{fig:agents}
\end{figure}
 
 Fig. \ref{fig:agents} illustrates the classes used to represent the agents, which are the decision-makers of the system. Our implementation includes two decision-makers, one for each type of event. The two agent classes are very similar, differing mainly in their action sets. The \texttt{FreeVehicleAgent} class is responsible for choosing a call from the pool of waiting calls, while the \texttt{NewCallAgent} class is responsible for choosing a vehicle. It is important to note that each of these has two deep q-networks. It is because we utilize the double-DQN algorithm to reduce the problem of bias maximization. The $b$ attribute in these classes refers to the hyperparameter of the reward function presented in the section \ref{sec:problem-formulation}. Most of the other parameters are related to the DQN algorithm and the neural networks involved.
 \begin{figure}
\centerline{\includegraphics[width=0.5\textwidth]{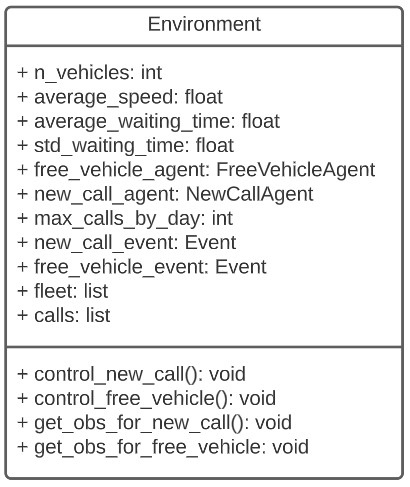}}
\caption{Environment}
\label{fig:environment}
\end{figure}

 The core of our simulator is the \texttt{Environment} class, which is presented in Figure \ref{fig:environment}. In addition to controlling the entire process flow, this class allows for the configuration of several simulation parameters, such as the number of vehicles in the fleet, the average speed of the vehicles, and the maximum number of daily calls. Additionally, the \texttt{Environment} class is responsible for providing state tuples to the agents through the \texttt{get\_obs\_for\_new\_call()} and \texttt{get\_obs\_for\_free\_vehicle()} methods.

\subsubsection{Event-handling routines}

A DES simulator works by maintaining a time-ordered list of events and processing the events by using specific event-handling routines. We describe below the two main routines in our simulator, which handle the \texttt{NewCallEvent} and \texttt{FreeVehicleEvent}.
\begin{figure}[t]
\centerline{\includegraphics[width=0.6\textwidth]{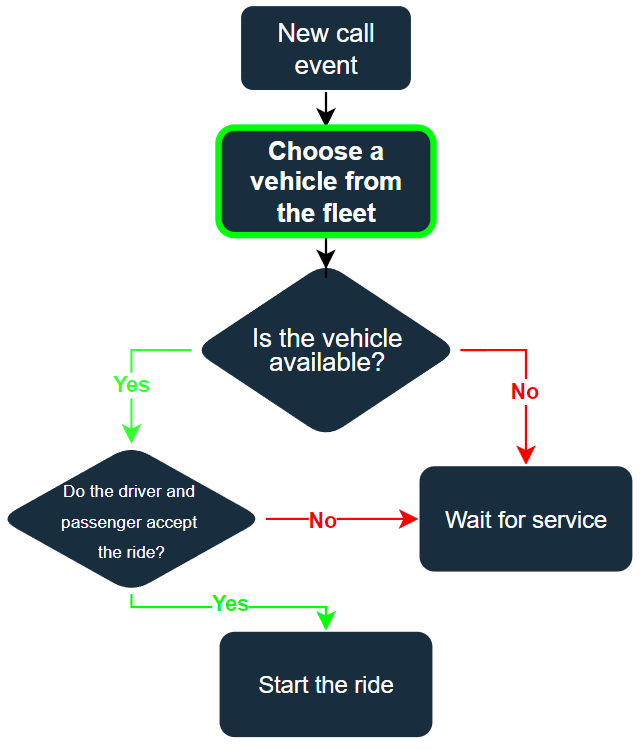}}
\caption{New call process flow}
\label{fig:new-call-flow}
\end{figure}

When a \texttt{NewCallEvent} is triggered, the \texttt{NewCallAgent} immediately scans all vehicles in the fleet to determine the most suitable one for the assignment. It then uses a decision policy to choose a particular vehicle. If the chosen vehicle is free, the agent initiates the assignment process, which requires confirmation from both the driver and the customer. On the other hand, if the selected vehicle is busy, the system places the call in a waiting call queue until it is assigned. It can occur, for instance, when a vehicle is finishing a trip near the call origin. Fig. \ref{fig:new-call-flow} shows the process flow of the routine, which handles the \texttt{NewCallEvent}. 

\begin{figure}[t]
\centerline{\includegraphics[width=0.8\textwidth]{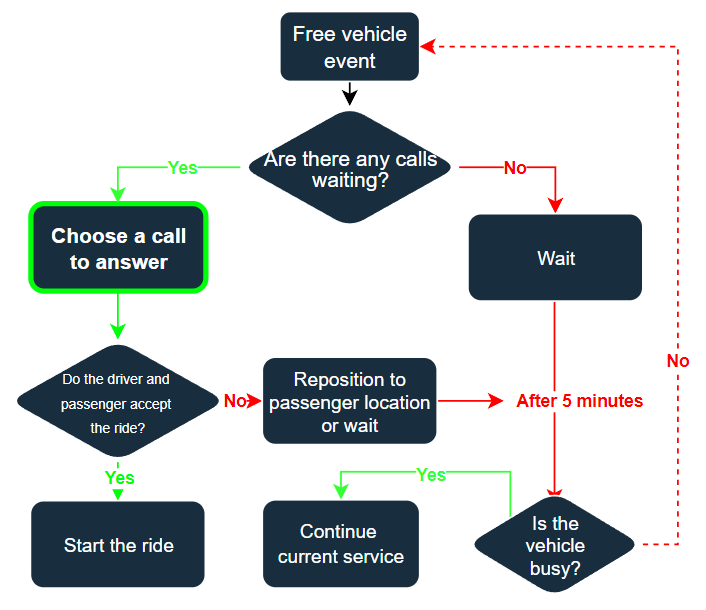}}
\caption{Free vehicle process flow}
\label{fig:free-vehicle-flow}
\end{figure}

When a \texttt{FreeVehicleEvent} is triggered, the \texttt{FreeVehicleAgent} queries the pool of waiting requests (calls). If there are calls waiting, it chooses one of the calls according to a decision policy. Otherwise, it does nothing and waits. Fig. \ref{fig:free-vehicle-flow} details the event-handling routines.

It should be noted that the \texttt{FreeVehicleAgent} will only need to decide if there is at least one call in the waiting calls pool. An important step in the flow is when the system suggests an assignment and must wait for the acceptance of both the driver and the customer. If both parties accept, the ride begins and the flow proceeds normally. However, in the occurrence of rejection by either party, the system will put the call to the waiting calls pool, and the driver will receive a recommendation from the \texttt{FreeVehicleAgent} to either reposition herself to the call location or await further instructions. In either scenario, the \texttt{FreeVehicleAgent} will wait for a maximum of 5 minutes until the vehicle is assigned to another call. If the vehicle remains unoccupied at the end of this period, a new \texttt{FreeVehicleEvent} will be triggered, and the flow will continue.

It is worth noting that the assignment of a vehicle to a call suggested by the agents is only a proposal for both the customer and the driver. Hence we consider the possibility of rejection by either party. To simulate the potential for rejection by either party, we employed two probability distributions in our modeling approach. The first distribution characterizes the probability of rejection by the driver, and it is modeled using a beta density function. For each vehicle, at its creation time, we sample a probability $p$ from the beta density, which represents the probability that the driver will reject an assignment. This implies that each vehicle is associated with a unique probability of rejecting an assignment, and when an agent attempts to make an assignment of the vehicle, a Bernoulli trial is performed using this probability. It is important to note that in a real-world application, this probability can be estimated by analyzing historical data on assignment rejections.

We employ an additional distribution that characterizes the customers' tolerance for waiting. Specifically, we adopt a gamma probability density to model this. During the call creation process, each call is assigned a maximum waiting time, which is randomly sampled from the gamma density. Consequently, when an agent proposes an assignment, customers will reject it if the estimated time for the vehicle to arrive at the pickup location exceeds the previously sampled maximum waiting time for the call.

\subsection{Agents' brains: deep neural networks}

Our intelligent agents are driven by deep neural networks which estimate q-values of current state and actions. In the case of the \texttt{NewCallAgent}, the action corresponds to assigning one of the available vehicles to the newly arrived call, while the action of the \texttt{FreeVehicleAgent} corresponds to assigning the free vehicle to one of the waiting calls. The agents decide on the best actions by querying their neural networks and identifying the action with the maximum q-value. Notice that as the agents perform one-to-many assignments, the number of possible actions is considerably smaller than when performing many-to-many assignments.
\begin{figure}
\centerline{\includegraphics[width=0.9\textwidth]{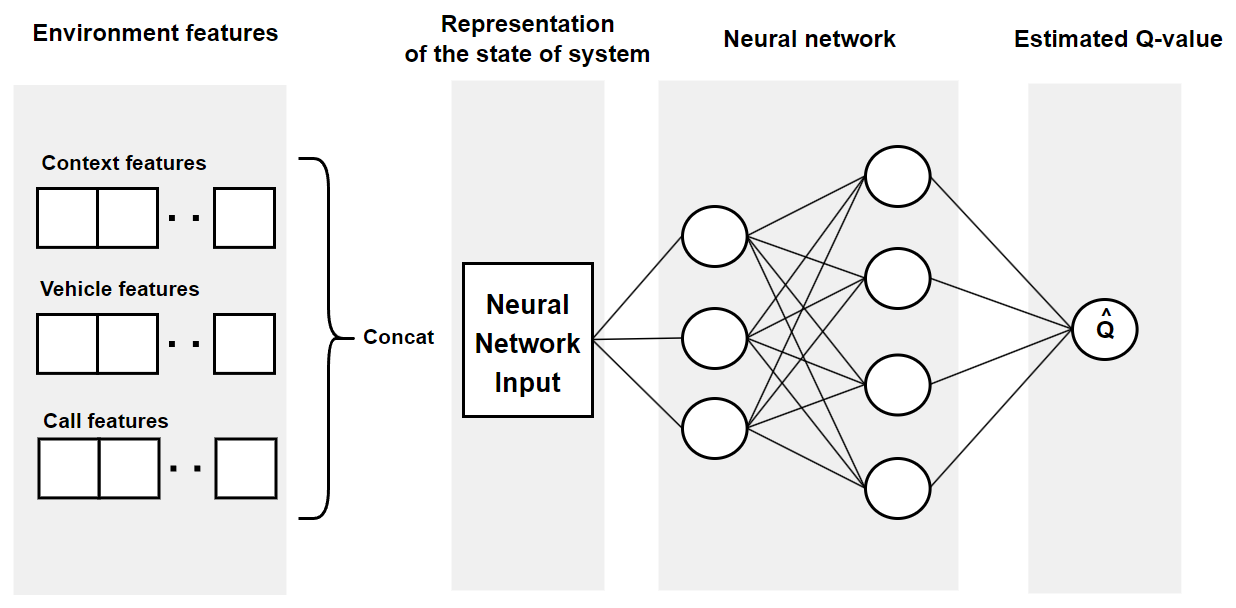}}
\caption{Data flow in the q-value estimation process}
\label{fig:state-representation}
\end{figure}

Fig. \ref{fig:state-representation} schematically shows the input and output of the deep neural networks. The input is the concatenation of the tuples presented in Section \ref{sec:problem-formulation} and the output it the corresponding estimate of the q-value, which represents the quality of the action. Each concatenated tuple represents a potential assignment between a vehicle and a call in a specific context. The agents receive several concatenated tuples and identify which one has the highest q-value. Notably, the list of tuples passed as input to the neural networks of the \texttt{FreeVehicleAgent} contains the same vehicle attributes and different call parameters, while for the \texttt{NewCallAgent}, the lists contain the same call attributes and different vehicle features.

To mitigate the overestimation bias inherent in traditional q-learning training \cite{watkinsLearningDelayedRewards1989}, we apply the double q-learning algorithm \cite{hasselt2010doubleqlearning}. This algorithm is an extension of the q-learning that addresses the overestimation issue caused by the maximum operator in its update formula, as shown in Eq. \eqref{eq:q-learning-update}. The proposed strategy in double q-learning is to introduce two distinct q-value functions, identified below by letters A and B, in the updating equation: 
\begin{equation}
    q^{A}(s,a)  \gets q^{A}(s,a)+\alpha \Big(r+e^{-\beta \tau}q^{B}(s', \argmax_{a' \in \mathcal{A}_{s'}}q^{A}(s', a')) - q^{A}(s,a)\Big).
    \label{eq:double-q-learning-update}
\end{equation}
Notice that, rather than using a single q-value function for both action selection and evaluation, double q-learning separates the functions, assigning one q-value function for action selection and another for action evaluation. 

Since double q-learning updates two different q-values simultaneously, in our implementation each of the two agents  have two deep neural networks dedicated to estimating the q-values, labeled network A and B. Then, given a sample transition $(s,a,s',r)$ at iteration $t$, the weights of network A are updated according to
\begin{equation}
\boldsymbol{\theta}^{\text{A}}_{t+1} \gets \boldsymbol{\theta}^{\text{A}}_t + \alpha (Y^\text{DoubleQ} - q(s,a;\boldsymbol{\theta}^{\text{A}}_t)) \nabla_{\boldsymbol{\theta}^{\text{A}}_t} q(s,a;\boldsymbol{\theta}^{\text{A}}_t),
\end{equation}
which is essentially a gradient descent step. The target is given by \cite{van2016deep}:
\begin{equation}
Y^\text{DoubleQ} = r + e^{-\beta \tau}q\Big(s', \argmax_{a' \in \mathcal{A}_{s'}}q(s', a'; \boldsymbol{\theta}^{\text{A}}_t); \boldsymbol{\theta}^{\text{B}}_t\Big),
    \label{eq:double-deep-q-learning-target}
\end{equation}
in which $\boldsymbol{\theta}^{\text{A}}_t$ denotes the current parameters of neural network A, while $\boldsymbol{\theta}^{\text{B}}_t$ represents the current parameters of neural network B. Network B is updated similarly, just swaping letters A and B in Eq. \eqref{eq:double-deep-q-learning-target}. 

The learning process begins after the EB has accumulated a certain number of samples (controlled by the parameter \texttt{learning\_starts}, shown in Fig. \ref{fig:agents}). We perform gradient steps using mini-batches sampled at each successful assignment in the environment. It is important to note that both neural networks have the same structure, except for their parameters. In one neural network, we perform gradient descent at every step, while in the other, we set the parameter \texttt{update\_steps} to control the number of steps before the network parameters are synchronized.

\subsection{Computational implementation}

Besides the abovementioned use of SimPy, we also use other Python libraries to implement the simulator and neural networks. Python 3.7.13 is utilized for the entire computational implementation of this work. Several packages are necessary, and the most important ones with corresponding versions are presented in the Table \ref{tab:libs}.
\begin{table}
	\centering
        \begin{tabularx}{0.4\hsize}{>{\hsize=.6\hsize}X>{\hsize=.4\hsize}X}
			\toprule
            
			Package & Version\\
			\midrule
			NumPy & 1.21.6\\
			Pandas & 1.3.5\\
			GeoPandas & 0.10.2 \\
			Simpy & 4.0.1\\
			Matplotlib & 3.5.1\\
			PyTorch & 1.8.2\\
            \bottomrule
		\end{tabularx}	
	\caption{\label{tab:libs} Python packages used in the computational implementation}	
\end{table}

\section{Numerical results}
\label{sec:results}

In our experiments, we utilized the for-hire vehicle trip records dataset provided by NYC Taxi and Limousine Commission\footnote{https://www.nyc.gov/site/tlc/about/tlc-trip-record-data.page} as input to the simulator. Specifically, we used the data from January 2022 for training our agents and February 2022 data to evaluate their performance. To restrict the scope of our study, we filtered the trips and included only those that occurred in Brooklyn and those completed through the Uber or Lyft platform. Distances were computed using the Manhattan distance.

\subsection{Training phase}

In the training phase, we set the discount factor for the agents to $\gamma = 0.9$ and the replay buffer size to 20,000. Additionally, we set the parameter \texttt{update\_steps} to synchronize the target and behavior networks at every 10,000 training steps. During training, we utilized an $\epsilon$-greedy policy with $\max\epsilon = 1$, $\min\epsilon = 0.05$, and an $\epsilon$-decrement of 0.99995. The agents started training once the replay buffer was half-full, which means that training started  when the replay buffer amounted to at least 10,000 experience tuples available.

The q-networks were constructed with two fully connected hidden layers and leaky-ReLU activation functions. The first hidden layer has 64 neurons, and the second has 32 neurons. The batch size is 32 and the learning rate is set to 0.001. We used Adam as optimization algorithm and used as loss function the smooth $\text{L}_1$ loss.

In order to train the agents to handle various situations, we defined four scenarios based on the ratio of fleet size in relation to the number of daily calls. For the very easy scenario, we set the fleet size to 3\% of the number of daily calls; for the easy scenario, we set it to 2\%; for the medium scenario, we set it to 1\%; and for the hard scenario, we set it to 0.5\%.
\begin{figure}[t]
\centerline{\includegraphics[width=0.9\textwidth]{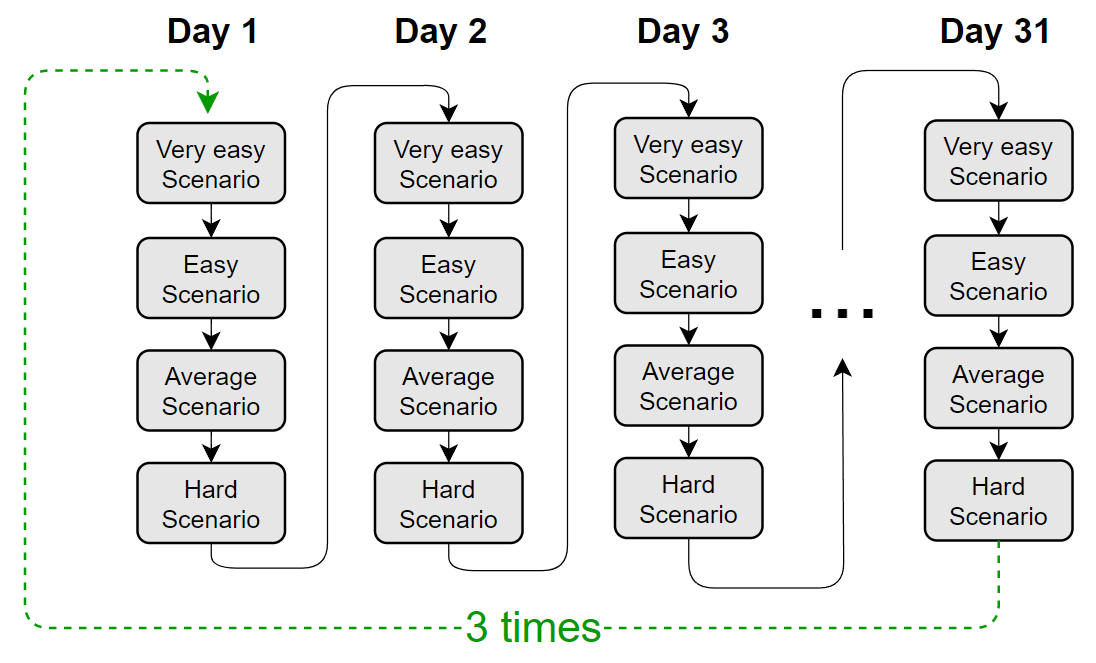}}
\caption{During the training phase, we iterate the 4 scenarios 3 times for each of the days of January 2022.}
\label{fig:training-phase}
\end{figure}

During the training phase, we set the maximum number of daily calls to 1,000 and we iterated 12 times on each day of the month, with 3 iterations for each scenario, as illustrated in Fig. \ref{fig:training-phase}. This leads to a total of 372 training days with each day having 24 hours. To evaluate the computational cost, we performed the training phase 10 times. Our findings showed an average duration of 34.63 minutes with a standard deviation of 1.05 minutes. The training was executed on a machine equipped with an Intel Core i9 13900k processor, 64 GB of memory running at 5600 MHz, and an RTX 4090 graphic card. During the training phase, the machine was solely dedicated to executing this task.

\begin{figure}
\centerline{\includegraphics[width=0.95\textwidth]{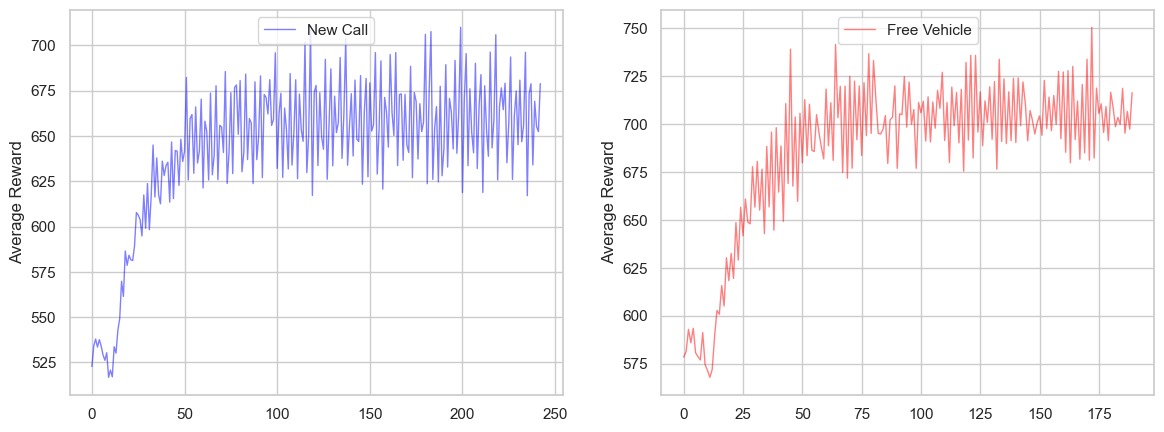}}
\caption{Average reward computed at intervals of 1,000 rewards received by each of the \texttt{NewCallAgent} and \texttt{FreeVehicleAgent} during the training phase}
\label{fig:training-rewards}
\end{figure}
\begin{figure}
\centerline{\includegraphics[width=0.95\textwidth]{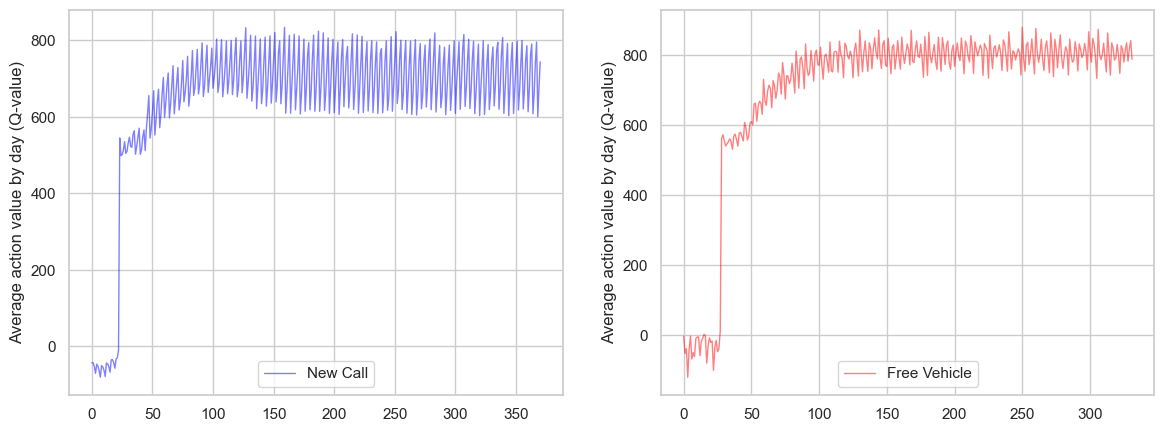}}
\caption{Average action values for every 1,000 chosen actions during the training phase}
\label{fig:training-q-values}
\end{figure}

\begin{figure}
\centerline{\includegraphics[width=0.95\textwidth]{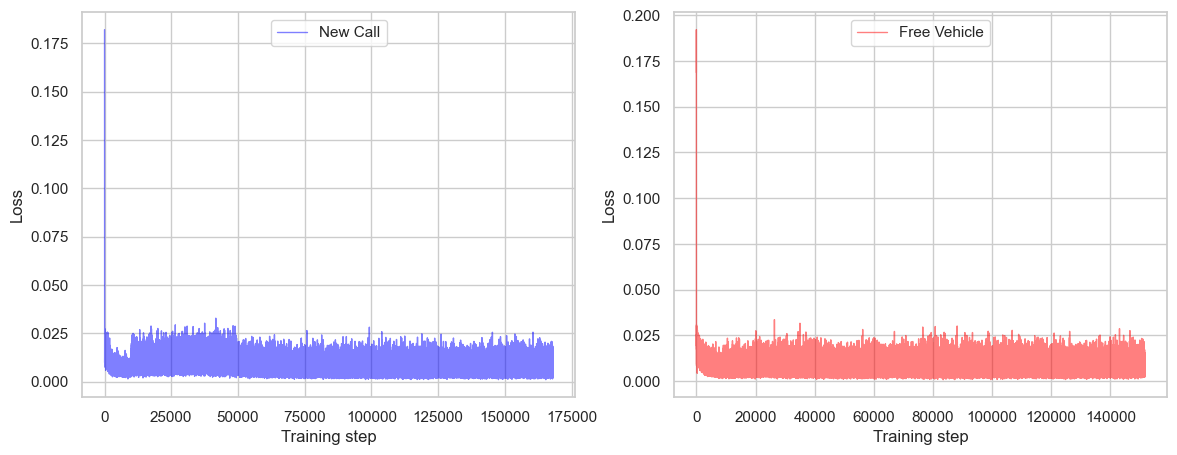}}
\caption{Loss function of the neural networks of both agents during the training phase}
\label{fig:training-loss-function}
\end{figure}

The average reward achieved during the training phase of the agents is presented in Fig. \ref{fig:training-rewards}. The average reward values shown were computed at intervals of every 1,000 rewards received. The evolution of average action values during the training phase for both agents is shown in Fig. \ref{fig:training-q-values}. These average values were computed for every 1,000 chosen actions. The convergence of neural networks is illustrated in Fig. \ref{fig:training-loss-function}, which show the decrease in the loss function. These plots demonstrate the rapid approximation of the value function by the neural networks during the training process.

\subsection{Testing phase}
During the testing phase, we set the number of daily calls to 100,000, surpassing the overall daily demand of Uber in Brooklyn. We conducted tests during entire month of February 2022. Notice that this data were kept fully separated from the data used during training, so that there is no data leakage and they were completely new to the trained agents.

We used as benchmarks to the trained policy four baseline policies, which are frequently used in practice due to their simplicity and easiness of implementation:
\begin{enumerate}
    \item First In First Out (FIFO): The agent always chooses the longest waiting call to assign a vehicle.
    \item Last In First Out (LIFO): The agent always chooses the most recent call to assign a vehicle.
    \item Nearest Neighbor (NN): The agent assigns the vehicle closest in distance to a new arriving call or the closest waiting call when a vehicle gets free. 
    \item Random: The agent chooses randomly a call or a vehicle at decision times.
\end{enumerate}

We compare policies according to three performance measures:
\begin{enumerate}
    \item Average delay: The average time taken by vehicles to reach the call origin. This time is measured from the moment the call is created until the vehicle arrives at the call origin.
    \item Cancellation rate: The ratio of canceled calls to the total number of calls received in a day.
    \item Total service time: The sum of the distances traveled from the origin to the destination for all completed trips in a day.
\end{enumerate}

\begin{figure}[t]
\centerline{\includegraphics[width=0.95\textwidth]{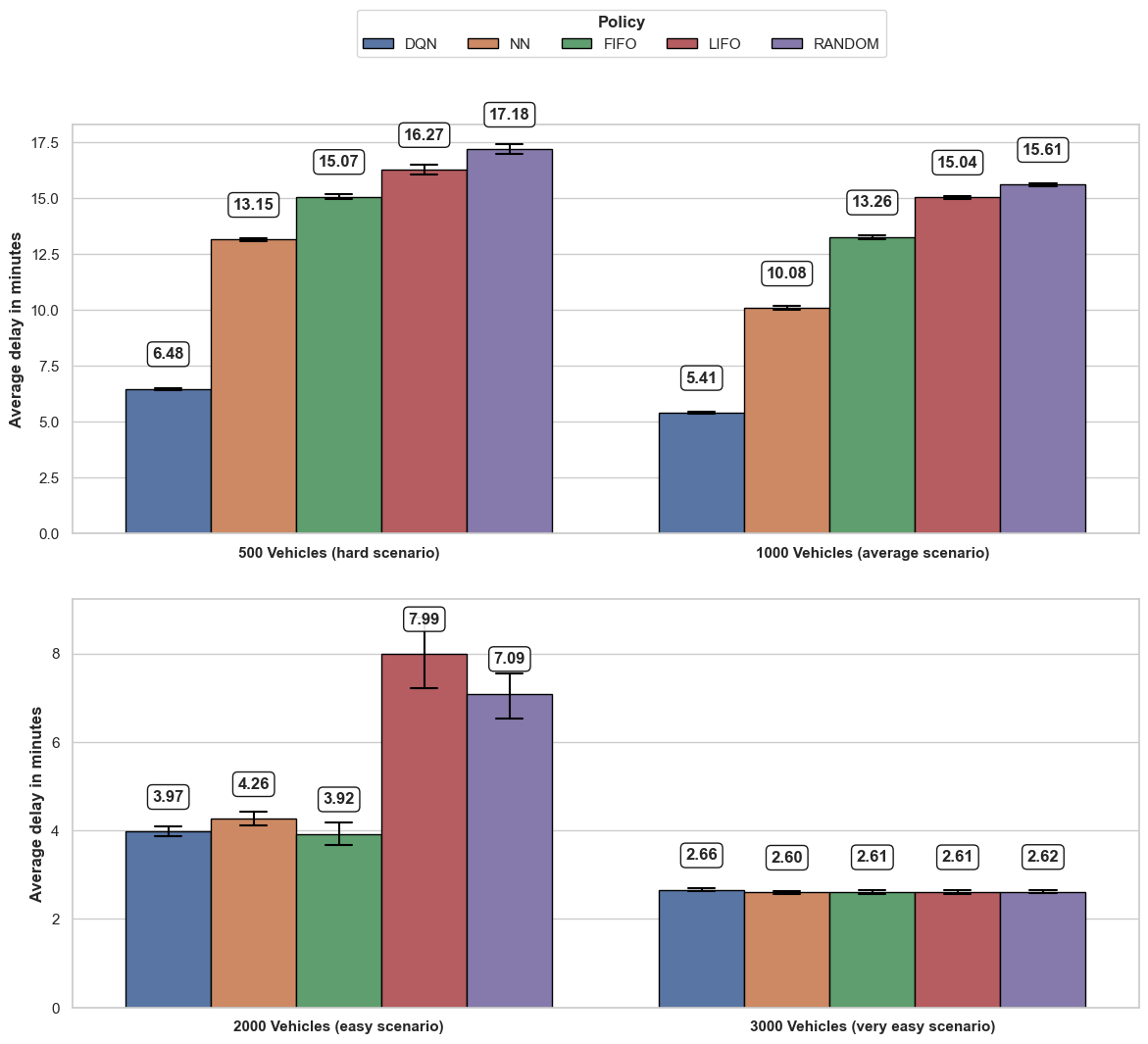}}
\caption{Average delay (in min.) during the test phase. Each bar represents the average daily delay over the days of February 2022. The horizontal lines at the top of bars represent 95\% confidence intervals.}
\label{fig:test-average-delay}
\end{figure}

\begin{figure}[t]
\centerline{\includegraphics[width=0.95\textwidth]{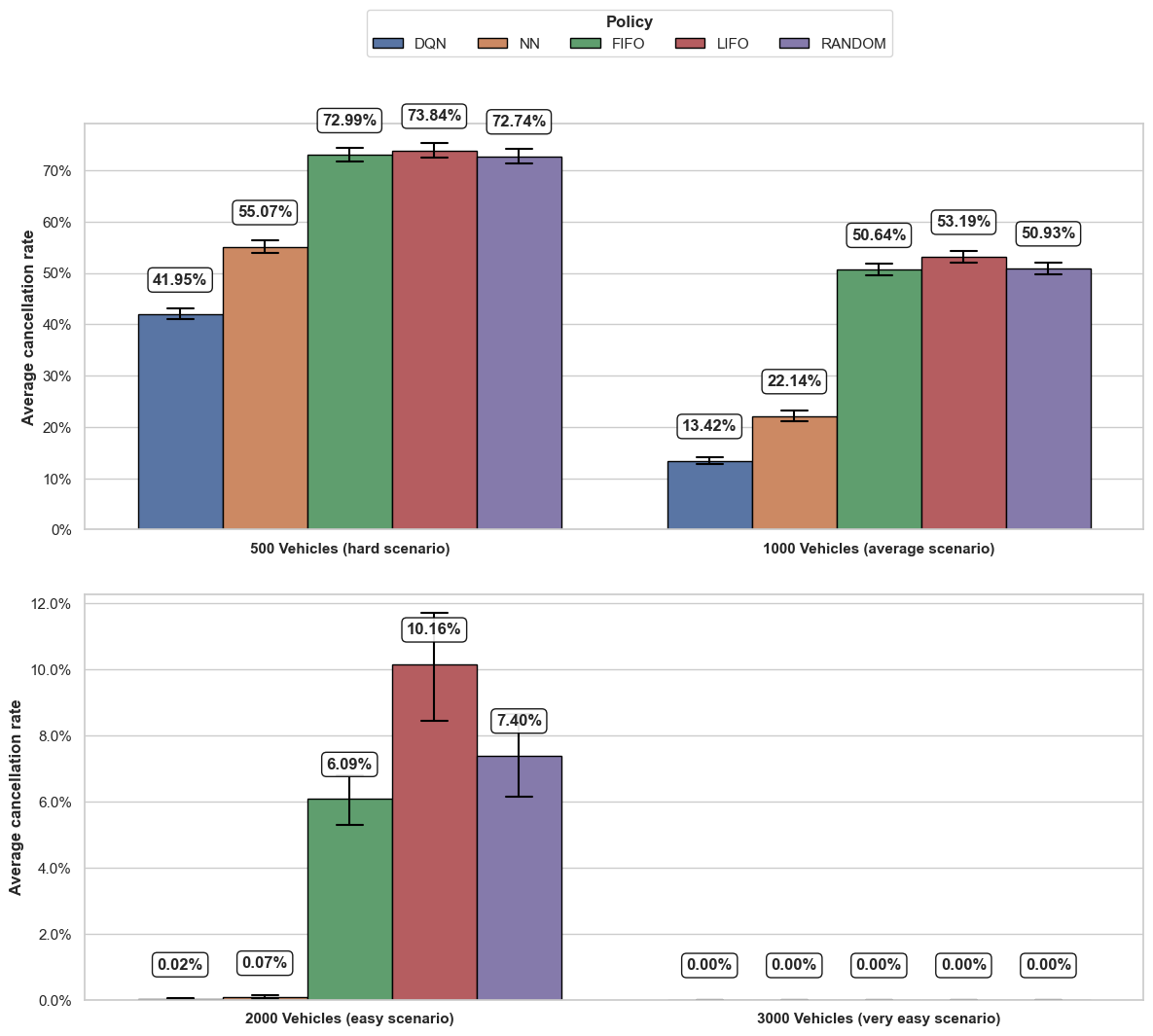}}
\caption{Cancellation rate during the test phase. Each bar represents the average daily cancellation rate over the days of February 2022. The horizontal lines at the top of bars represent 95\% confidence intervals.}
\label{fig:test-cancellation-rate}
\end{figure}

\begin{figure}[t]
\centerline{\includegraphics[width=0.94\textwidth]{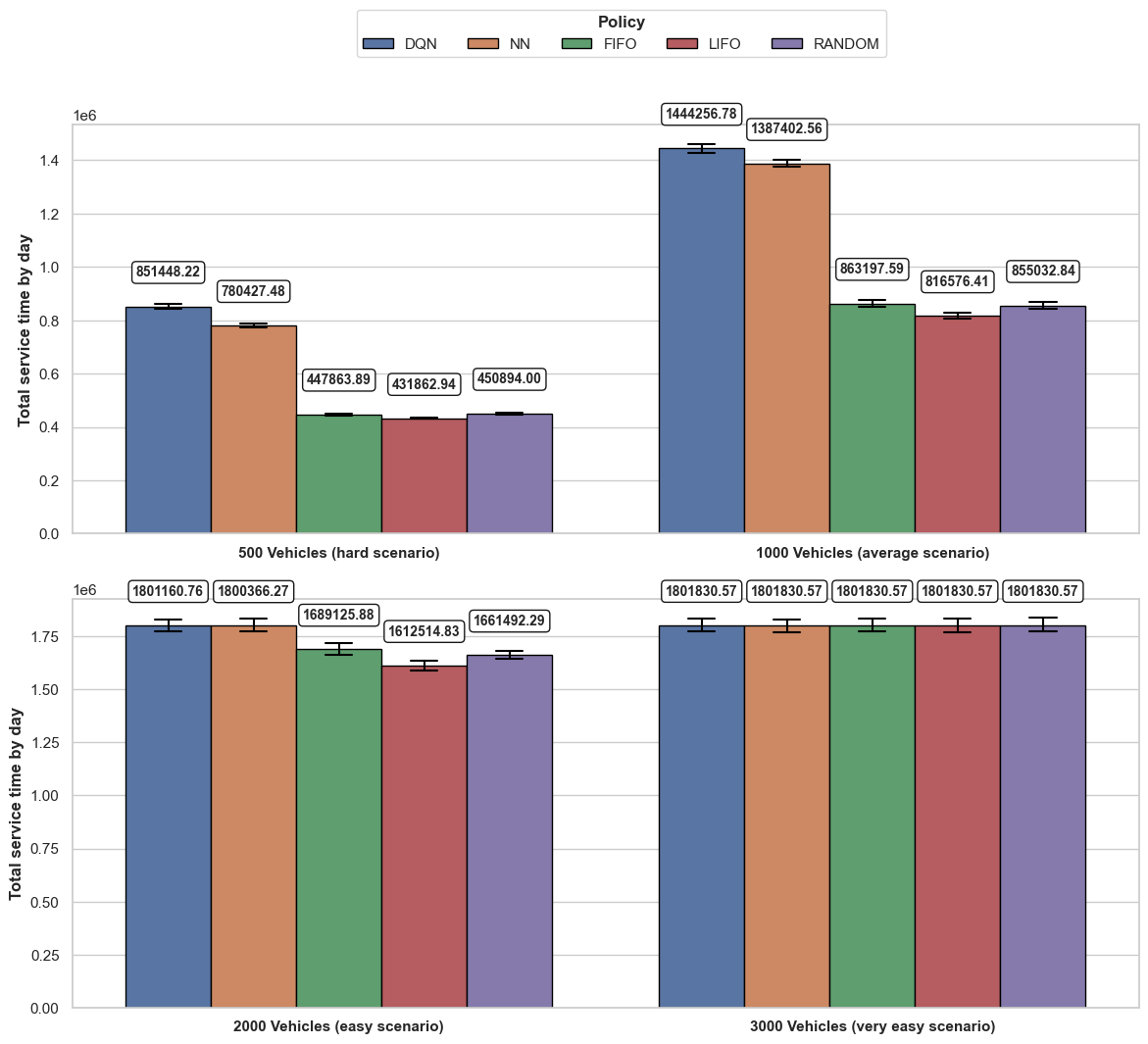}}
\caption{Total service time during the test phase. Each bar represents the average daily service time over the days of February 2022. The horizontal lines at the top of bars represent 95\% confidence intervals.}
\label{fig:test-total-service-time}
\end{figure}

Fig. \ref{fig:test-average-delay} presents the average daily delay of our agents, referred to as DQN, along with the baseline policies, during the testing phase. It is important to note that none of the policies had prior exposure to the demand in this scenario. Fig. \ref{fig:test-cancellation-rate} illustrates the average daily cancellation rate while Fig. \ref{fig:test-total-service-time} displays the average daily total service time of all policies.

\section{Discussion}
\label{sec:discussion}

The plots shown in Figs. \ref{fig:training-rewards}, \ref{fig:training-q-values}, and \ref{fig:training-loss-function} demonstrate the fast convergence of our training method. The average q-values exhibit consistently low variance over time, especially for the free vehicle event. On the other hand, the average rewards of the agent responsible for this event display some instability during training, but in the long term the average rewards become acceptable. Additionally, it is noticeable that both the rewards and q-values do not exhibit significant increases after a certain point in the training phase, suggesting that a shorter training duration may be sufficient to obtain results with similar performance measures.

Based on the findings depicted in Fig. \ref{fig:test-average-delay}, a clear advantage of our agents over the baseline policies becomes evident in the challenging and moderately challenging scenarios. In the hard scenario we achieved a reduction of approximately 50\%  and in the average scenario a reduction of 46\%  in average delay when compared with the second best policy (NN). However, this advantage is not so prominent in the less demanding scenarios. Specifically, in the easy scenario, the delay achieved by the DQN approach is minimal, yet the NN and FIFO policies also demonstrate competitive performance in this regard. Furthermore, these three policies achieved narrow confidence intervals, indicating consistent results. In the scenario characterized as ``very easy'', wherein the number of drivers significantly surpasses the demand, all policies exhibit comparable performance. This outcome arises from the substantial pool of available drivers in this scenario, resulting in extensive spatial coverage of the map and minimization of the probability of a lack of nearby vehicles for new ride requests.

Fig. \ref{fig:test-cancellation-rate} also provides evidence of the superiority of the policy learned by our agents compared to the alternative policies. Notably, in the hard and average scenarios, our approach achieved a significant reduction of approximately 13.12\% and 8.72\%, respectively, in the cancellation rate. This reduction can have a profound impact on mitigating overall customer dissatisfaction, indicating the effectiveness of our learned policy in minimizing cancellations. In the easy scenario, the NN policy exhibits exceptional performance, comparable to our approach. It is also worth noticing that, as can be seen in Fig. \ref{fig:test-average-delay}, the FIFO policy exhibits comparable performance to our DQN policy. On the other hand, in Fig. \ref{fig:test-cancellation-rate} we can see that FIFO exhibits higher cancellation rate than our DQN policy. As an aside, in the very easy scenario, all policies achieve perfect performance due to the abundant availability of vehicles, as in the previous analysis of average delay.

Finally, we consider the total service time performance measure. Total service time is a proxy for total revenue of drivers, since the more time drivers are servicing customers the higher their revenues. As illustrated in Figure \ref{fig:test-total-service-time}, our approach consistently surpasses all baseline policies in both the hard and average scenarios, corroborating the trends observed in previously analyzed performance measures. In these scenarios, the NN policy emerges as the second-best alternative, demonstrating its efficacy as a myopic policy. In the easy and very easy scenarios, all policies exhibit similar performance, with the LIFO policy being the least favorable. Notably, in the easy scenario, our approach and the NN policy achieve comparable performance, resulting in a technical draw. In the very easy scenario, all policies demonstrate flawless performance, due to the absence of cancellations, as evidenced by Figure \ref{fig:test-cancellation-rate}. Consequently, maximum total service time is achieved across all policies in the very easy scenario.

\section{Conclusions}
\label{sec:conclusions}

In this paper, we proposed a new event-based approach for the dynamic vehicle dispatching
problem. We formulated the problem as a semi-Markov decision process and developed a solution approach by integrating a discrete-event simulator with deep reinforcement learning. To this end, we employed the double deep q-learning algorithm to train two different agents corresponding to new call and free car events. Agents' policies were approximated by deep neural networks. Our developed discrete-event simulator incorporates often overlooked characteristics in existing literature, such as the probability of a driver rejecting an assignment proposed by the system and the maximum waiting time tolerance of customers. We carried out extensive experiments that closely mirror real-world conditions with the use of real data from the city of New York and compared with alternative heuristic policies often used in practice.

The decision policy obtained through our proposed approach achieved a substantial reduction of up to 50\% in the average customer waiting time relative to the second-best policy corresponding to assigning the nearest vehicle. Moreover, our approach also resulted in the lowest average cancellation rates and highest total service time, a proxy for total revenues received by drivers. Notice that achieving low cancellation rates is critical, since it mitigates potential customer disappointments, discouraging customers from seeking alternative services and promoting customer loyalty. Additionally, the increase in total service time contributes to elevated driver satisfaction within the mobility platform, since it results in more revenue and less idle time, leading to a more productive and fulfilling experience for them.

In future works, we aim to extend our approach by employing alternative RL algorithms and comparing their performance against recently proposed matching strategies. Additionally, we plan to enhance the capabilities of our simulator by incorporating real-time traffic information. Finally, we intend to explore the potential of multi-agent approaches.

\section*{Acknowledgments}
Funding: This study was financed in part by Conselho Nacional de Desenvolvimento Científico e
Tecnológico (CNPq; Grant No.: 407466/2021-5). We also thank NVIDIA Corporation for the GPU support.

\bibliographystyle{unsrtnat}
\bibliography{references}

\end{document}